\documentclass[article,10pt]{elsarticle} 
\usepackage{titlesec}
\usepackage{float}
\usepackage[T1]{fontenc}
\usepackage{amsmath}
\usepackage{svg}
\usepackage{pdfpages}
\usepackage{bm}
\usepackage{tikz}
\usetikzlibrary{positioning,shapes.geometric}
\usepackage[unicode=true, bookmarks=true]{hyperref}
\usepackage[utf8]{inputenc}
\usepackage{times} 
\usepackage[verbose,tmargin=1in,bmargin=1in,lmargin=1in,rmargin=1in]{geometry}
\usepackage{setspace} 
\usepackage{mdframed}
\usepackage{geometry}
\usepackage{booktabs}
\usepackage{multirow}
\usepackage{multicol, makecell}
\usepackage{tabularx}
\usepackage{longtable}
\usepackage{threeparttable} 
\usepackage{array}
\usepackage{url}
\usepackage{parskip}
\usepackage{setspace}
\usepackage[utf8]{inputenc}
\usepackage{graphicx}

\newmdenv[
  linecolor=black,
  linewidth=0.5pt,
  backgroundcolor=gray!10,
  roundcorner=5pt,
  skipabove=10pt,
  skipbelow=10pt,
  innerleftmargin=10pt,
  innerrightmargin=10pt,
  innertopmargin=10pt,
  innerbottommargin=10pt
]{contextbox}
\usepackage[superscript]{cite}
\usepackage{xcolor}
\usepackage{titlesec}
\newcommand{\absdiv}[1]{%
  \par\addvspace{.5\baselineskip}%
  \noindent\textbf{#1}\quad\ignorespaces}

\makeatother
\journal{Nature Machine Intelligence}

\graphicspath{{figures/}}

\begin{document}
\begin{frontmatter}

\title{Few-shot Cross-country Generalization of Tabular Machine Learning and Foundation Models for Childhood Anemia Prediction under Distribution Shift}


\author[1]{Yusuf Brima\corref{cor1}}
\cortext[cor1]{Corresponding author: ybrima@uos.de}
\author[2,3]{Marcellin Atemkeng} 
\author[4]{Lansana Hassim Kallon}
\author[5]{David Niyukuri}
\author[6] {Antoine Vacavant}
\author[7]{Samuel Saidu}
\author[8,9]{Ding-Geng Chen}

\address[1]{Computer Vision, Institute of Cognitive Science, Osnabr\"{u}ck University, Germany}
\address[2]{Department of Mathematics, Rhodes University, South Africa}
\address[3]{National Institute for Theoretical and computational Sciences (NITheCS), Stellenbosch, 7600, South Africa}
\address[4]{DAHW - German Leprosy \& Tuberculosis Relief Association, Mano River Cluster (Sierra Leone \& Liberia)}
\address[5]{Interdisciplinary Research Program in Public Health, University of Burundi, Burundi}
\address[6]{Universite Clermont Auvergne, Clermont Auvergne INP, CNRS, Institut Pascal, Clermont–Ferrand, France}
\address[7]{Department of International Public Health, Liverpool School of Tropical Medicine, Liverpool, UK}
\address[8]{College of Health Solutions, Arizona State University, Phoenix, USA}
\address[9]{Department of Statistics, University of Pretoria, Pretoria, South Africa}

\begin{abstract}
\absdiv{\textcolor{purple}{Background}}

Childhood Anemia affects an estimated 40\% of children aged 6--59 months globally and arises from heterogeneous nutritional, infectious, and socioeconomic factors that vary substantially across settings. This variability challenges the generalizability of predictive machine learning models, which often degrade under cross-population or temporal shifts. We investigated the utility a modern transformer-based tabular foundation model (TabPFN) as a complementatry framework with respect to supervised classical machine learning methods across diverse country contexts, with particular attention to data-scarce settings where surveillance capacity is most limited.

\absdiv{\textcolor{purple}{Methods}}

We conducted a multi-country prediction study using Demographic and Health Surveys (DHS) children's recode data from 16 countries spanning Africa, Asia, Latin America, the Caucasus, and the Middle East. The harmonized analytic cohort comprised of $(n=68,856)$ children aged 6--59 months with valid hemoglobin measurements. Anemia was defined using WHO age and altitude-adjusted thresholds and treated as a binary outcome. We trained Logistic Regression, XGBoost, and LightGBM models using standard supervised learning, and evaluated TabPFN v2.6 in an in-context learning setting. Performance was assessed using Area Under the Receiver Operating Characteristic Curve (AUC-ROC) and other standard classification metrics, with calibration evaluated via Brier score and expected calibration error (ECE). Uncertainty in performance estimates was quantified using bootstrap resampling to derive 95\% confidence intervals. Robustness was assessed in a few-shot learning setting. Cross-population generalization was examined using leave-one-country-out (LOCO) validation and reverse-LOCO experiments to assess directional transferability. Subgroup analyses were conducted across five demographic strata: child age group, sex, maternal education, residence type, and household wealth quintile. Feature importance was assessed using standard linear and tree-based explainer SHAP values for the three supervised models and an adapted version of SHAP for TabPFN, aggregated across countries and examined at the country level.

\absdiv{\textcolor{purple}{Findings}}

In data-scarce regimes, TabPFN consistently outperformed supervised models: below approximately 200 within-country training samples, it achieved higher discrimination than logistic regression, XGBoost, and LightGBM, whose performance degraded more rapidly under data scarcity. TabPFN also yielded the best probabilistic calibration across all 16 countries, achieving the lowest mean Brier score (0.203) and Expected Calibration Error (ECE = 0.042) of all models evaluated; LightGBM and Logistic Regression exhibited the greatest miscalibration, particularly at higher predicted probabilities. Under full-data conditions, within-country discrimination was moderate across all models (AUC-ROC 0.59--0.76) and varied substantially by country, with consistently higher performance in Armenia and Guatemala and lower performance in Liberia and Sierra Leone; differences between models within the same country were small (typically $\leq$0.05 AUC points) and confidence intervals overlapped substantially within each country. Under LOCO validation, performance declined modestly (AUC-ROC 0.58--0.69) and remained primarily driven by the target country rather than model choice. Reverse-LOCO analyses revealed asymmetric and directional transferability, with epidemiologically diverse populations serving as more informative training sources and certain target populations remaining persistently difficult to predict regardless of model or training data. Subgroup analyses across 13 countries with sufficient data showed broadly similar AUC-ROC values across demographic strata within any given country, with no model exhibiting consistent bias toward specific subgroups; the dominant source of variation was country context rather than model architecture. Feature importance analysis with SHAP identified child age as the most influential predictor across all four models and all countries, followed consistently by altitude and height-for-age z-score. In the upper tier, wealth index and maternal education ranked prominently, while illness-related and supplementation variables contributed modestly and with greater cross-country variability.

\absdiv{\textcolor{purple}{Interpretation}}

Predictive performance in childhood Anemia modeling is currently constrained more by cross-population heterogeneity and data characteristics than by model architecture. Within this constraint, transformer-based tabular foundation models provide a meaningful and practically relevant advantage when labeled data are limited, offering both improved discrimination and better-calibrated risk estimates. Subgroup and interpretability analyses reinforce that performance ceilings are set by epidemiological and data-structural factors rather than by differential model behavior across demographic groups or predictor sets. These properties are strongest precisely in the low-resource surveillance contexts where prediction tools are most urgently needed. These findings shift emphasis from model selection toward addressing population-level structural challenges, while establishing foundation models as a promising pathway for Anemia prediction in data-scarce global health settings.
\end{abstract}

\begin{keyword}
childhood Anemia \sep global health \sep in-context learning \sep tabular machine models \sep few-shot learning \sep tabular foundation models
\end{keyword}

\end{frontmatter}
\section{Introduction}
\label{sec:intro}
Childhood Anemia remains a major public health problem worldwide and is especially concentrated in settings with high burdens of under-nutrition, infection, and socioeconomic disadvantage. The World Health Organization (WHO) estimates that 40\% of children aged 6--59 months are affected globally, with the highest burden in the African and South-East Asian regions~\cite{who_anemia_2025}. In the Global Burden of Disease Study 2021, Anemia remained a substantial cause of nonfatal health loss, with 52.0 million years lived with disability worldwide in 2021 and the greatest burden among young children, women, and populations in sub-Saharan Africa~\cite{gardner2023prevalence}. Beyond its prevalence, childhood Anemia is clinically important because it is associated with impaired cognitive and motor development, reduced growth and activity, and broader downstream effects on human capital~\cite{who_anemia_2025,gardner2023prevalence}.

The causes of childhood Anemia are multi-faceted and vary substantially across geographic settings. Nutritional deficiencies, infection, inflammation, maternal health, household conditions, and broader socioeconomic factors can all contribute, and their relative importance differs by geography and population context~\cite{who_anemia_2025,tesfaye2024application}. This stark heterogeneity makes prediction difficult and helps explain why models developed in one country may not transfer well to another. It also underscores the need for methods that can account for nonlinear relationships and distribution shift across settings.

The Demographic and Health Surveys (DHS) program offers a crucial foundation for this type of work. DHS surveys are nationally representative and use standardized questionnaires and data structures across countries, which makes them well suited to comparative and cross-country analysis~\cite{dhs_dataset_types}. The children's recode (KR) data, for each country, contains one record for each child born to interviewed women in the five years preceding the survey, enabling harmonized child-level prediction studies across countries and survey rounds~\cite{dhs_dataset_types}. This structure allows models to be developed and evaluated under more realistic transport settings than those used in many single-country studies.

There has been a grounds well of interest and an increasingly growing body of literature has applied machine learning to childhood Anemia prediction, predominately binary classification of Anemia vs. non Anemia, but most studies have remained country specific. In Bangladesh, a DHS-based study analyzed 2,013 children and reported that Random Forest achieved an accuracy of 68.53\% and an AUC of 0.6857, while Logistic Regression reached an accuracy of 62.75\% and an AUC of 0.6276~\cite{khan2019machine}. In Ethiopia, a study using the 2016 DHS data and a weighted sample of 8,482 children reported that Logistic Regression achieved an accuracy of 66\% and an AUC of 0.69, whereas Random Forest achieved an accuracy of 64\% and an AUC of 0.63~\cite{tesfaye2024application}. More recent work in Assam, India, analyzed 8,411 children from National Family Health Survey (NFHS-5) and reported that Gradient Boosting achieved an accuracy of 77.2\% and an AUC of 0.83 after balancing the data with Synthetic Minority Over-sampling Technique for Nominal and Continuous features (SMOTE-NC)~\cite{das2025optimizing}. In Nigeria, a study of 13,136 children from the 2018 National DHS data found that the Extra Trees classifier achieved an AUC of 0.8319 and an accuracy of 0.7565~\cite{ja2025predicting}. In Ghana, a study of 3,382 children from the 2022 DHS data reported that Gradient Boosting achieved an AUC of 0.72, with an accuracy of 66.27\% and an F1 score of 68.99\%~\cite{hassan2026predicting}. Taken together, this body of work suggests that machine learning can identify useful predictors, but it leaves open the question of how well these models transfer across epidemiologically varied populations. A further underexplored area is that performance estimates are overwhelmingly reported as point values without uncertainty quantification, making it difficult to assess whether observed differences between models are reliable or within the range of sampling variability.

Quite recently, the upsurge in tabular foundation model research offers a paradigm shift. Tabular Prior-data Fitted Network (TabPFN) was introduced as a transformer-based model that performs, amongst other things, downstream supervised prediction through \textit{in-context learning in a single forward pass}, without task-specific training~\cite{hollmann2025tabpfn}. Subsequent work on TabPFN v2 showed improved handling of heterogeneous tabular data and strong in-context learning performance across a wide range of downstream tasks~\cite{ye2025closer,brima2025robustness}. However, its use in global child health prediction, particularly under cross-country distributional variability, remains underexplored. This creates an opportunity to evaluate whether such a \textit{generalist model} adds value over more established supervised approaches in a public health setting where labeled data paucity but heterogeneity are prevailing challenges.

In this work, we use DHS children's recode data from 16 countries spanning Africa, Asia, the Caribbean, the Middle East, and the Caucasus to predict childhood Anemia under within-country, (LOCO), reverse LOCO, and few-shot learning settings. We investigated TabPFN v2.6 in tandem with Logistic Regression, XGBoost, and LightGBM, complementing discrimination across varied dimensions and calibration analyses with subgroup evaluation and feature attribution methods. In doing so, we make three principal contributions. First, we provide one of the most geographically diverse multi-country studies of childhood Anemia prediction, using a harmonized analytic pipeline that enables cross-country comparison. Second, we systematically characterize the limits of cross-population generalization through LOCO and reverse LOCO experiments, revealing systematic patterns in cross-population knowledge transfer. Third, we demonstrate the practical value of in-context learning utilized by transformer-based tabular foundation models especially in data-scarce public health settings, showing that TabPFN provides both improved discrimination and better-calibrated risk estimates when labeled data are scarce, a regime that characterizes many low-resource surveillance contexts. Across our analyses, performance estimates are accompanied by 95\% confidence intervals derived from bootstrap resampling, enabling a more principled assessment of inter-model differences and the reliability of reported results.

The remainder of this paper is organized as follows. Section~\ref{sec:methods} describes the data sources, cohort derivation, outcome definition, predictors, and analytical methods. Section~\ref{sec:results} presents within-country, cross-country, few-shot, calibration, subgroup, decision curve, and interpretability results. Section~\ref{sec:discussion} synthesizes and contextualizes the findings, and Section~\ref{sec:conclusion} draws conclusions and outlines directions for future work.

\section{Methods}
\label{sec:methods}
\subsection{Study design and reporting framework}
We conducted a multi-country predictive modeling study using retrospective DHS data between 2008 to 2024. The goal was to develop and evaluate models predicting Anemia status under internal and external validation scenarios. The analysis followed TRIPOD-AI guidelines for transparent model development and validation~\cite{collins2024tripodplus}. The study emphasizes predictive accuracy rather than causal inference.

\subsection{Data source and study population}
We used DHS KR datasets from 16 countries (see Table~\ref{tab:dhs_summary}), covering surveys within a 16 year period across Africa, Asia, Latin America, Caucasus, and the Middle East. DHS surveys are nationally representative population-based surveys with standardized questionnaires, conducted under uniform protocols. The KR data contains one record per child born to a sampled woman in the last five years~\cite{demographic2017health,dhs_using_datasets}. We curated data from each country and applied a harmonized preprocessing pipeline. Eligible subjects were children age 6--59 months with measured hemoglobin and complete data for prespecified predictors. We excluded records with missing hemoglobin values or many missing covariates. After filtering, the final analytic sample included 68{,}856 children (see Table~\ref{tab:dhs_summary} for country-wise sample sizes and characteristics).

\begin{table}[!ht]
\centering
\begin{tabular}{l r r r r r r r}
\toprule
Country & Raw $N$ & Final $N$ & Age (mo) & Male (R) & Female (R) & Male (F) & Female (F) \\
\midrule
Gabon 2019--2021 & 6376 & 4462 & 32.10 & 3296 & 3080 & 2297 & 2165 \\
Malawi 2024 & 12312 & 3923 & 33.30 & 6152 & 6160 & 1953 & 1970 \\
Rwanda 2019--2020 & 8092 & 3303 & 32.74 & 4095 & 3997 & 1670 & 1633 \\
Sierra Leone 2019 & 9899 & 3367 & 31.53 & 5047 & 4852 & 1720 & 1647 \\
Haiti 2016-2017 & 6530 & 4773 & 32.88 & 3299 & 3231 & 2411 & 2362 \\
Jordan 2023 & 9106 & 4209 & 35.03 & 4739 & 4367 & 2146 & 2063 \\
Armenia 2015-2016 & 1724 & 1313 & 33.25 & 931 & 793 & 707 & 606 \\
Kyrgyz Republic 2012 & 4363 & 3437 & 32.00 & 2244 & 2119 & 1754 & 1683 \\
Liberia 2019--2020 & 5704 & 2048 & 31.87 & 2825 & 2879 & 997 & 1051 \\
Bolivia 2008 & 8605 & 2233 & 33.25 & 4430 & 4175 & 1179 & 1054 \\
Congo DRC 2021--2022 & 24000 & 9002 & 33.07 & 12208 & 11792 & 4593 & 4409 \\
Guatemala 2014--2015 & 12440 & 9987 & 33.39 & 6445 & 5995 & 5134 & 4853 \\
Ghana 2022 & 9353 & 3742 & 31.54 & 4804 & 4549 & 1901 & 1841 \\
Gambia 2019--2020 & 8362 & 3146 & 32.34 & 4364 & 3998 & 1650 & 1496 \\
Cameroon 2018 & 9733 & 3825 & 32.46 & 4938 & 4795 & 1928 & 1897 \\
Mali 2023--2024 & 15631 & 6086 & 33.06 & 7883 & 7748 & 3104 & 2982 \\
\midrule
\textbf{Total/Mean} &  152230 & 68856 & 32.85 & 77700 & 74530 & 35144 & 33712 \\
\bottomrule
\end{tabular}
\caption{Characteristics of DHS child cohorts across the 16 study countries. Raw $N$ refers to the original KR sample; final $N$ refers to the analytic cohort after preprocessing. (R = raw sample, F = final analytic sample).}
\label{tab:dhs_summary}
\end{table}
\subsection{Outcome definition}
The binary outcome was Anemia status, derived from the DHS hemoglobin indicator (\texttt{hw57}). DHS codebooks define four Anemia categories (none, mild, moderate, severe) based on WHO age and altitude-adjusted hemoglobin thresholds~\cite{who_anemia_indicator,who_hemoglobin_guideline_2024,world2024guideline}. For children aged 6--59 months, Anemia classification is defined as: no Anemia (hemoglobin $\geq 11.0$ g/dL), mild Anemia (10.0--10.9 g/dL), moderate Anemia (7.0--9.9 g/dL), and severe Anemia ($<7.0$ g/dL). We collapsed mild, moderate, and severe Anemia (hemoglobin $<11.0$ g/dL) into a single ``Anemia'' category and compared this against ``no Anemia''. This binary definition supports robust modeling across diverse settings while aligning with clinical significance.

\subsection{Predictors}
We selected predictors based on prior literature and availability across surveys~\cite{yimer2025optimizing,ja2025predicting,said2025hybrid,hassan2026predicting}. Child-level variables included age (months), sex, anthropometric z-scores (height-for-age, weight-for-height, weight-for-age), recent illnesses (fever, diarrhea, cough), deworming treatment, and vitamin A supplementation. Maternal variables included age, age at first birth, education level, and smoking status. Household socioeconomic factors included wealth index quintile, urban/rural residence, primary water source, sanitation facility, household structure, and parity. Geographic altitude (for hemoglobin adjustment) was also included. Detailed variable names and codes are given in Table~\ref{tab:features}. We applied a consistent cross-country harmonization: categorical variables were turned into sparse categorical encodings, and continuous measures were standardized using parameters from training data.

\begin{table}[!ht]
\centering
\caption{Feature domains, DHS variables, and descriptions used in Anemia prediction models}
\label{tab:features}
\small
\begin{tabular}{p{3cm} p{4.2cm} p{7.5cm}}
\toprule
\textbf{Domain} & \textbf{DHS Variables} & \textbf{Description} \\
\midrule

Child &
\texttt{hw1, b4, hw70, hw72, h22, h11, h31, h43, h34} &
Age (months), sex, anthropometric z-scores, recent illness (fever, diarrhea, cough), deworming status, vitamin A supplementation \\[6pt]

Maternal &
\texttt{v012, v212, v106, v463z} &
Maternal age, age at first birth, education level, smoking status \\[6pt]

Household SES &
\texttt{v190, v191, v025, v113, v116, v151, v201} &
Wealth index, residence type, water source, sanitation, household structure, parity \\[6pt]

Geographic &
\texttt{v040} &
Altitude used for hemoglobin adjustment \\[6pt]

Outcome &
\texttt{hw57} &
WHO-defined Anemia status \\

\bottomrule
\end{tabular}
\end{table}

\subsection{Missing data and preprocessing}
Records with missing outcome data (\texttt{hw57}) were excluded prior to analysis, and the sample was restricted to children aged 6--59 months; full details of cohort derivation are provided in Appendix (Figure~\ref{fig:cohort_derivation_flowchart}; Table~\ref{tab:exclusion_counts}). Missing predictor values, present in 36,704 records (53.3\% of the analytic sample) and were handled by imputation: median imputation was applied to continuous predictors and mode imputation to categorical predictors. Imputation parameters were derived exclusively from training data and applied to validation and test folds to prevent data leakage. Continuous predictors were standardized using z-scores computed from the training set. Categorical variables were encoded numerically prior to model training. All preprocessing steps were fitted on training data only and applied to held-out folds, ensuring no leakage across splits.

\subsection{Analytical methods}
We evaluated four modeling approaches. Logistic Regression with L2 regularization~\cite{nunez2011regression} served as an interpretable linear model. XGBoost~\cite{chen2016xgboost} and LightGBM~\cite{ke2017lightgbm} represented the gradient-boosted tree family, with LightGBM additionally employs histogram binning, leaf-wise tree growth, and gradient-based one-side sampling to accelerate training. TabPFN v2.6~\cite{hollmann2025tabpfn} is a transformer-based tabular foundation model that performs in-context learning without gradient updates, generating predictions by conditioning on a support set of labeled examples at inference time. We evaluated TabPFN in both a full in-context learning setting and a few-shot setting in which a small fraction of labeled samples from the target country were included in the support set. For logistic regression, XGBoost, and LightGBM, hyperparameters were optimized using Optuna's Bayesian optimization~\cite{akiba2019optuna} within cross-validation on the training data (see~\ref{app:hyperparameter_search}); TabPFN required no additional training, and its performance depends solely on the choice of in-context demonstration examples at inference.

\subsection{Validation strategy}
We performed both internal and external validation. Internal validation used stratified 5-fold cross-validation within each country, with models trained and tested on disjoint folds of the same country's data while preserving Anemia prevalence across splits. External validation proceeded in two complementary directions. In the standard LOCO framework, for each country $k$ we trained on the pooled data from all remaining countries and evaluated on country $k$, simulating deployment of a model to a previously unseen population. In the reverse LOCO framework, each model was instead trained on a single source country and evaluated on each of the remaining 15. These experiments enabled assessment of the directional transferability of country-specific predictive signal and identification of which populations generalize best as training source. Together, these two external validation schemes characterize both the practical utility of multi-country pooling and the nature of cross-population knowledge transfer. To assess model behavior under data scarcity, we conducted a separate within-country few-shot analysis in which each country's analytic sample was partitioned into a small training set of varying size of country-specific records and a held-out test set comprising the remainder; supervised models were retrained from scratch at each sample size, while for TabPFN the selected samples constituted the in-context support set at inference, with no gradient-based updating, so that its few-shot curve reflects how in-context learning scales with support set size rather than training set size in the conventional sense.

\subsection{Performance measures}
Our primary metric was the area under the receiver operating characteristic curve (AUC-ROC), which summarizes discrimination. We also report accuracy, precision, recall, and F1-score. Since Anemia prevalence varied across countries, Macro F1 helps evaluate performance under class imbalance. Calibration was assessed with the Brier score (mean squared error of predicted probabilities) and the Expected Calibration Error (ECE) (the weighted difference between confidence and accuracy across probability bins). Lower Brier and ECE indicate better calibration~\cite{breiman2001statistical}. To quantify uncertainty in all performance estimates and support more principled comparison between models, we derived 95\% confidence intervals for each metric using bootstrap resampling with 5{,}000 iterations, sampling with replacement from the held-out test.

\subsection{Model interpretability}
To understand feature effects, we used SHapley Additive exPlanations (SHAP) values for model interpretability across all evaluated models~\cite{lundberg2017unified}. SHAP provides a unified framework that assigns each feature an importance value for individual predictions based on cooperative game theory~\cite{lundberg2017unified}. For the tree-based models (LightGBM and XGBoost), SHAP values were computed using the standard TreeSHAP implementation.

For TabPFN, we used the shapiq framework, which provides Shapley-based explanations specifically adapted for in-context learning models such as TabPFN through a remove-and-recontextualize interpretation paradigm~\cite{muschalik2024shapiq,rundel2024interpretableTabPFN}. This approach enables the estimation of feature attributions while accounting for the contextual inference mechanism underlying TabPFN predictions. Global feature importance rankings and local explanation analyses were generated from the resulting SHAP values.

All interpretability analyses were descriptive rather than causal and were intended to compare the relative contribution of input variables across models.

\subsection{Experimental Setup}
All analyses were implemented in Python using scikit-learn, XGBoost, LightGBM, Optuna, and the TabPFN library. The Code for this work is publicly available. We ran analyses on a system with NVIDIA A100 Graphics Processing Unit (GPU) for TabPFN; training of tree models took minutes per model.

\subsection{Ethical considerations}
This was a secondary analysis of anonymized, publicly available DHS data. The original DHS surveys obtained ethical approval from respective local review boards and complied with informed consent~\cite{dhs_using_datasets,demographic2017health}. Approval for this analysis secondary use was granted by the DHS Program.

\section{Results}
\label{sec:results}
We present results across eight complementary analyses: few-shot learning, model calibration, within-country discrimination, cross-country generalization in a LOCO setting, reverse LOCO, subgroup performance, decision curve analysis, and finally feature interpretability. We lead with the few-shot and calibration analyses because these reveal intriguing and practically meaningful differences between models. The remaining analyses characterize the broader landscape of predictive performance and its determinants across the 16 study populations.
\subsection{Few-shot scenario analysis}
\label{subsec:fewshot}
\begin{figure}[!htp]
\centering
\includegraphics[width=\textwidth]{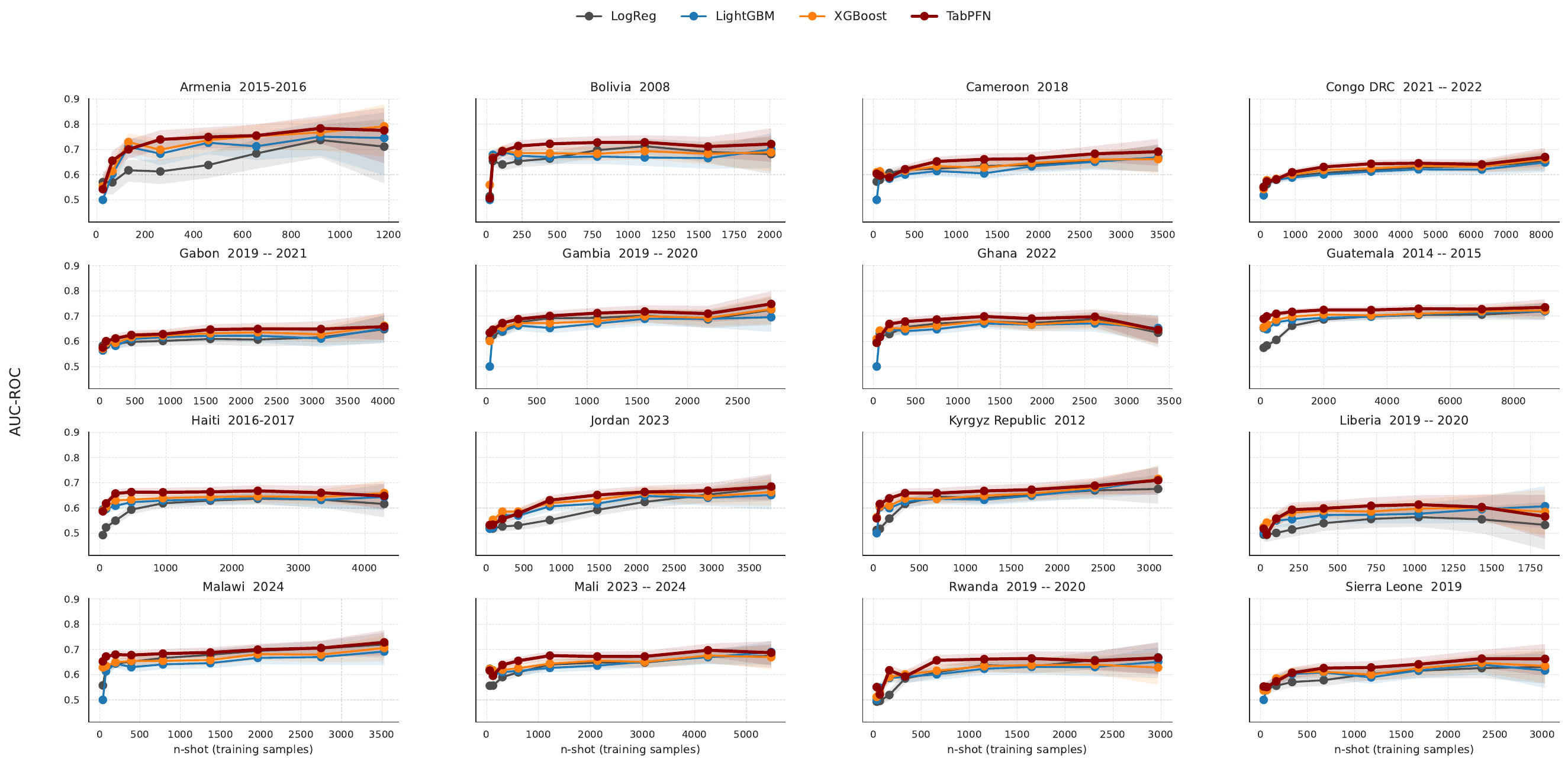}
\caption{\textbf{Few-shot performance curves (AUC-ROC) by country and model as a function of within-country training sample size.} Each panel shows AUC-ROC on the held-out test set as a function of the number of within-country labeled training samples ($n$-shot) for all models. The $x$-axis scale differs across panels, reflecting variation in country-specific analytic sample sizes. TabPFN leverages in-context learning and does not undergo gradient-based parameter updates; its $n$-shot curve reflects increasing in-context support set size. All supervised models were retrained from scratch at each sample size. Shaded regions indicate 95\% confidence intervals derived from bootstrap resampling across random sub-sampling replicates.}
\label{fig:fewshot_auc_by_country}
\end{figure}
Performance curves across all 16 countries showed, unsurprisingly, that AUC-ROC improved monotonically with increasing numbers of within-country training samples for all four models, though the rate and ceiling of improvement varied substantially by country (Figure~\ref{fig:fewshot_auc_by_country}). At very low sample sizes (fewer than approximately 200 within-country training observations), TabPFN achieved consistently higher AUC-ROC values relative to the three supervised models, which showed steeper degradation under extreme data scarcity, with non-overlapping bootstrap confidence intervals at the lowest sample sizes supporting the reliability of this advantage. This pattern was most pronounced in Guatemala, Haiti, Malawi and Gambia, where TabPFN's AUC-ROC in the near-zero training regime  exceeded that of logistic regression by approximately 0.10--0.15 units. As within-country labeled data accumulated, the advantage of TabPFN over gradient-boosted models narrowed considerably, and XGBoost or LightGBM reached comparable discrimination at larger sample sizes, with confidence intervals overlapping substantially in higher-resource regimes. Logistic regression showed the most consistent sensitivity to sample size, starting lowest at near-zero-shot across most countries and improving most steeply with additional training data, ultimately converging toward gradient-boosted model performance in higher-resource regimes. The relative ordering of models was most variable at low sample sizes, where confidence intervals were also widest, reflecting greater uncertainty in performance estimates under extreme data scarcity; ordering stabilized as $n$-shot increased and intervals narrowed. Country-level variation in performance curves was notable: in settings with larger total sample sizes and higher Anemia prevalence (e.g.\ Congo DRC, Guatemala, Mali), AUC-ROC continued to improve across the full range of observed training sizes, suggesting that additional labeled data remained informative. In smaller or lower-prevalence settings (e.g.\ Armenia, Liberia), the curves plateaued at lower $n$-shot values, consistent with an inherent ceiling imposed by the difficulty of the prediction task or limited signal in available predictors.

\subsection{Calibration and probabilistic reliability}
\label{subsec:calibration}
\begin{figure}[!htp]
\centering
\includegraphics[width=0.85\textwidth]{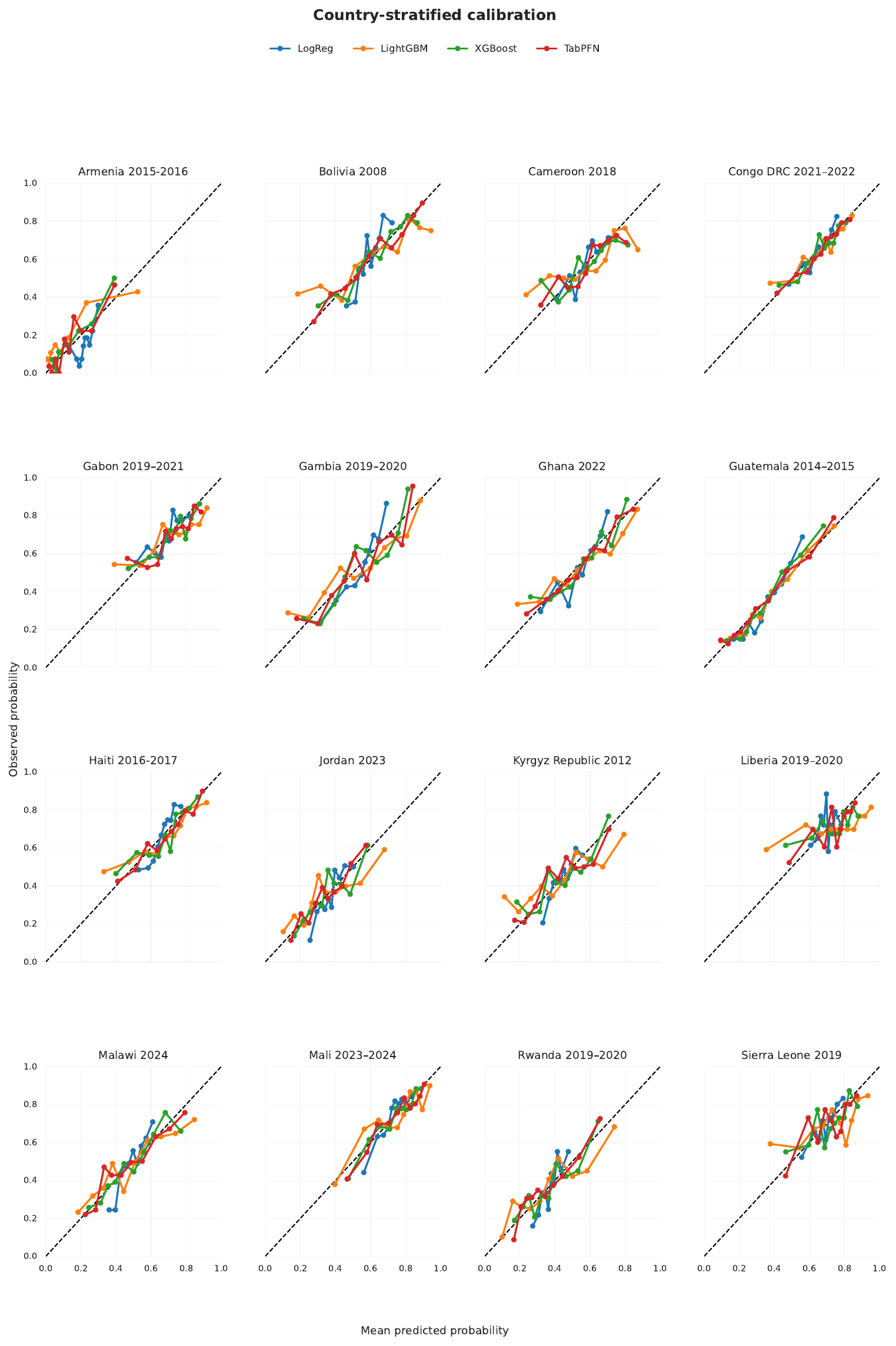}
\caption{\textbf{Country-stratified calibration curves for all four predictive models.} Each panel shows observed Anemia frequency (y-axis) against mean predicted probability (x-axis) across ten equal-width probability bins for each model. The dashed diagonal represents perfect calibration. Curves above the diagonal indicate underestimation of risk; curves below indicate overestimation. Calibration was assessed where each model was trained on the train split for each country and evaluated on the held-out test set for that country.}
\label{fig:figure_calibration_country_clean}
\end{figure}

Beyond discrimination, TabPFN produced the best-calibrated probability estimates across countries, achieving the lowest mean Brier score (0.203) and mean ECE (0.042) of all four models (Table~\ref{tab:calibration_summary}). XGBoost ranked second on both metrics (Brier: 0.207; ECE: 0.048), followed by logistic regression (Brier: 0.212; ECE: 0.052), with LightGBM showing the largest miscalibration (Brier: 0.213; ECE: 0.069). Calibration curves across all countries showed that predicted probabilities from all four models were broadly aligned with observed Anemia frequencies, with most curves tracking reasonably close to the diagonal reference line (Figure~\ref{fig:figure_calibration_country_clean}). However, the degree of alignment varied both across countries and across models. LightGBM exhibited the most visible miscalibration, with its curve frequently deviating from the diagonal particularly in the upper probability range, consistent with its higher mean ECE. Logistic regression showed a tendency toward modest overconfidence in some settings and under-confidence in others. XGBoost and TabPFN produced curves generally closer to the diagonal, with TabPFN showing somewhat more consistent alignment across the probability range, particularly in countries with moderate Anemia prevalence such as Congo DRC, Guatemala, and Haiti. Calibration was noticeably less stable in countries with smaller analytic samples or more extreme Anemia prevalence: in Armenia, all models showed irregular calibration curves with wider deviations from the diagonal, likely reflecting limited sample size constraining reliable probability estimation. Jordan and Kyrgyz Republic similarly showed greater between-model divergence, with some models systematically overestimating predicted probabilities at the upper end of the scale. These patterns are detailed in the country-level Brier score and ECE values reported in Appendix Table~\ref{tab:country_calibration}.

\begin{table}[!ht]
\centering
\caption{Across-country model calibration performance. Lower values indicate better
probabilistic calibration and predictive reliability across validation countries.}
\label{tab:calibration_summary}
\begin{threeparttable}
\begin{tabular}{lccccc}
\toprule
\textbf{Model} & \textbf{Mean Brier $\downarrow$} & \textbf{SD Brier} & \textbf{Mean ECE $\downarrow$} & \textbf{SD ECE} & \textbf{Rank (Brier) $\downarrow$} \\
\midrule
TabPFN   & 0.203 & 0.026 & 0.042 & 0.014 & 1 \\
XGBoost  & 0.207 & 0.027 & 0.048 & 0.014 & 2 \\
LogReg   & 0.212 & 0.024 & 0.052 & 0.015 & 3 \\
LightGBM & 0.213 & 0.028 & 0.069 & 0.022 & 4 \\
\bottomrule
\end{tabular}
\begin{tablenotes}
\footnotesize
\item $\downarrow$ indicates lower values are better. ECE = Expected Calibration Error.
SD = standard deviation across countries. Brier score measures probabilistic accuracy;
lower is better.
\end{tablenotes}
\end{threeparttable}
\end{table}

\subsection{Within-country model performance}
\label{subsec:within_country}

\begin{figure}[!htp]
\centering
\includegraphics[width=\textwidth]{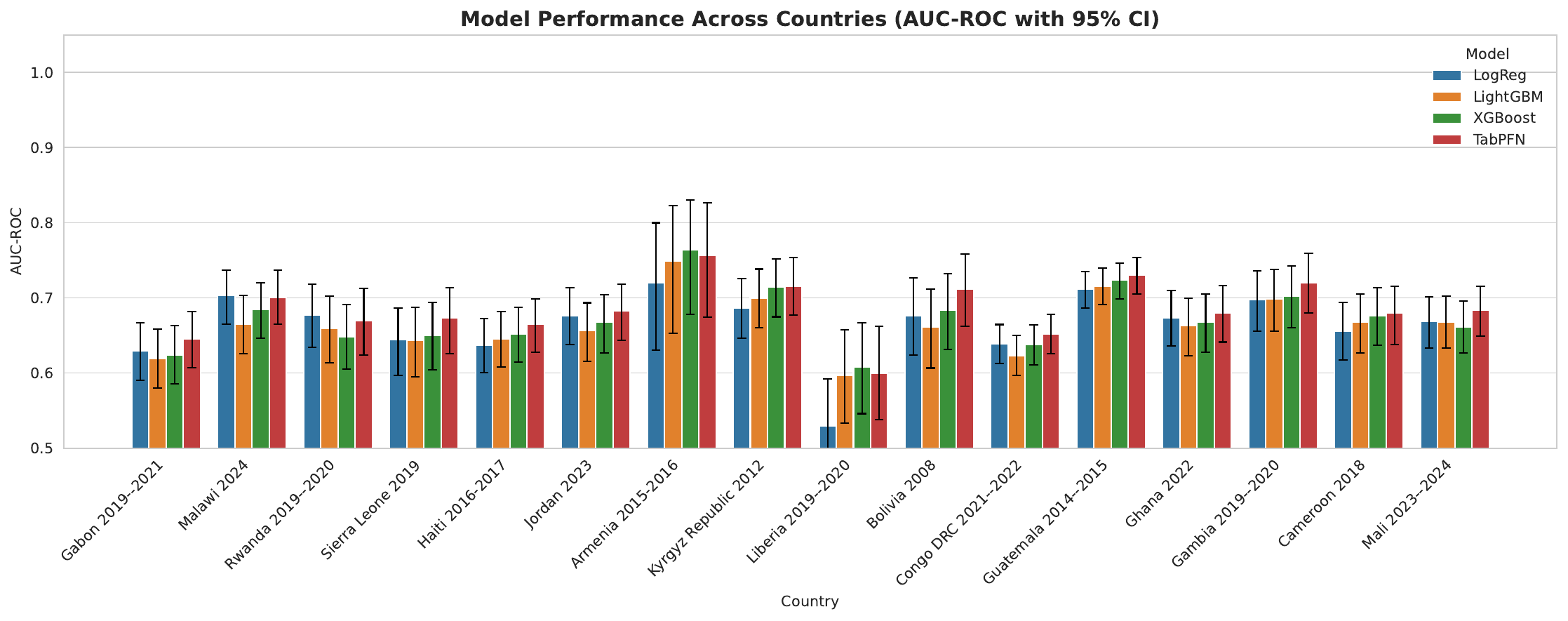}
\caption{\textbf{Within-country discriminative performance (AUC-ROC) of four predictive models across 16 study populations.} AUC-ROC from stratified five-fold cross-validation within each country for the models, with error bars denoting 95\% bootstrap confidence intervals. Differences between models within any single country are modest and confidence intervals overlap substantially; variation across countries exceeds variation across models, reflecting heterogeneity in Anemia epidemiology and data characteristics across study settings.}
\label{fig:anemia_model_performance_auc}
\end{figure}

Within each country, all four models achieved moderate discriminative performance under stratified five-fold cross-validation, with AUC-ROC values ranging from approximately 0.53 to 0.76 across the 16 study populations (see Figure~\ref{fig:anemia_model_performance_auc} and Appendix Table~\ref{tab:auc_results}). The highest discrimination occurred in Armenia and Guatemala, where all models attained AUC-ROC values between 0.72 and 0.76, while the lowest occurred in Liberia and Sierra Leone, where point estimates ranged from 0.53 to 0.67 irrespective of model. Bootstrap confidence intervals were notably wider in smaller samples such as Armenia and Liberia, reflecting greater uncertainty in performance estimates in those settings, and intervals overlapped substantially across models within every country. Within any given country, gradient-boosted trees and TabPFN tended to yield marginally higher discrimination than logistic regression, but these differences were consistently small, with the gap between the best and worst model within a country rarely exceeding 0.02--0.05 AUC points, and confidence intervals overlapping in all cases. TabPFN achieved the highest point estimate in 11 of 16 countries, though the magnitude of its advantage over the next-best model was modest and confidence intervals overlapped in all cases. LightGBM and XGBoost tracked closely throughout. The dominant source of performance variation was the country itself rather than model architecture: cross-country differences in AUC-ROC substantially exceeded inter-model differences at every study site, suggesting that epidemiological heterogeneity, data characteristics, and survey-level factors set the effective ceiling on within-country predictive performance.

\subsection{Cross-country generalization (LOCO)}
\label{subsec:loco}
\begin{figure}[!ht]
\centering
\includegraphics[width=\textwidth]{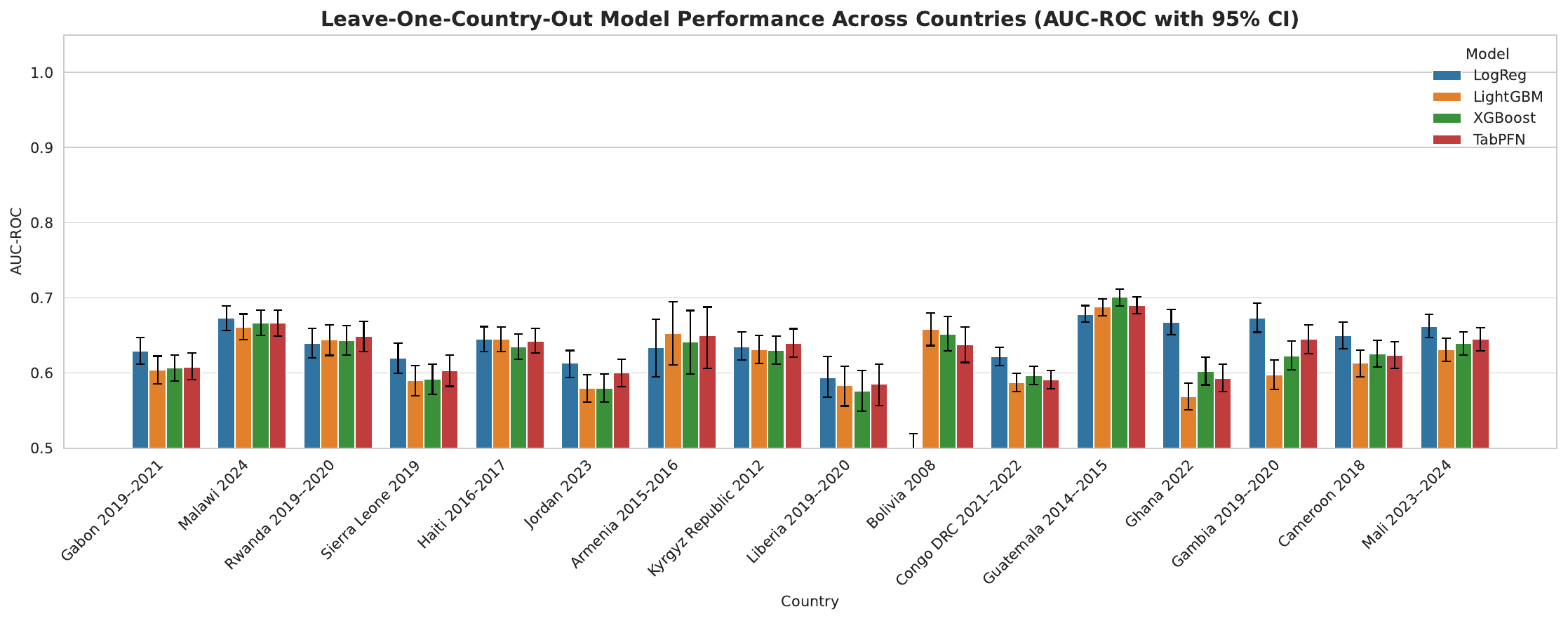}
\caption{\textbf{External discriminative performance (AUC-ROC) under LOCO validation across 16 study populations.} Each country was iteratively held out as an external test set while models were trained on the pooled data from the remaining 15 countries. AUC-ROC is shown for each model with error bars denoting 95\% bootstrap confidence intervals. Performance declines relative to within-country cross-validation across all models, reflecting distributional shift between training and target populations. Confidence intervals overlap substantially across models within any given held-out country, and inter-country variation in AUC-ROC exceeds inter-model variation throughout.}
\label{fig:anemia_loco_auc}
\end{figure}

Under LOCO validation, discriminative performance declined modestly but consistently relative to within-country cross-validation across all models. Held-out AUC-ROC values spanned 0.58 to 0.69, with most country-level estimates clustering between 0.62 and 0.66 (Figure~\ref{fig:anemia_loco_auc} and Appendix Table~\ref{tab:loco_auc_results}), indicating that moderate discrimination persisted on average under distributional shift. Bootstrap confidence intervals were generally wider under LOCO than under within-country cross-validation, reflecting the added uncertainty introduced by distributional shift between training and target populations. Guatemala yielded the highest external AUC-ROC across all models (0.68--0.69), consistent with its large analytic sample size and nutritional risk profile proximate to the pooled training distribution. Performance was lowest when Ghana or Liberia was held out, with AUC-ROC point estimates falling below 0.62 for most models, suggesting that the remaining training countries provided limited transferable signal to these populations; confidence intervals in these settings were also among the widest, further cautioning against strong conclusions about model behavior in these specific target populations. Inter-model differences remained small and inconsistent: logistic regression matched or exceeded tree-based models in several countries (notably Ghana and Gambia), while XGBoost and TabPFN performed strongest in others (notably Guatemala, Armenia, and Malawi), with margins rarely exceeding 0.03--0.04 AUC points and confidence intervals overlapping in all cases. As in the within-country analysis, cross-country variability in AUC-ROC substantially exceeded inter-model variability, reinforcing that the identity of the target population was a stronger determinant of external predictive performance than model choice.

\subsection{Reverse LOCO analysis}
\label{subsec:reverse_loco}
\begin{figure}[!ht]
\centering
\includegraphics[width=\textwidth]{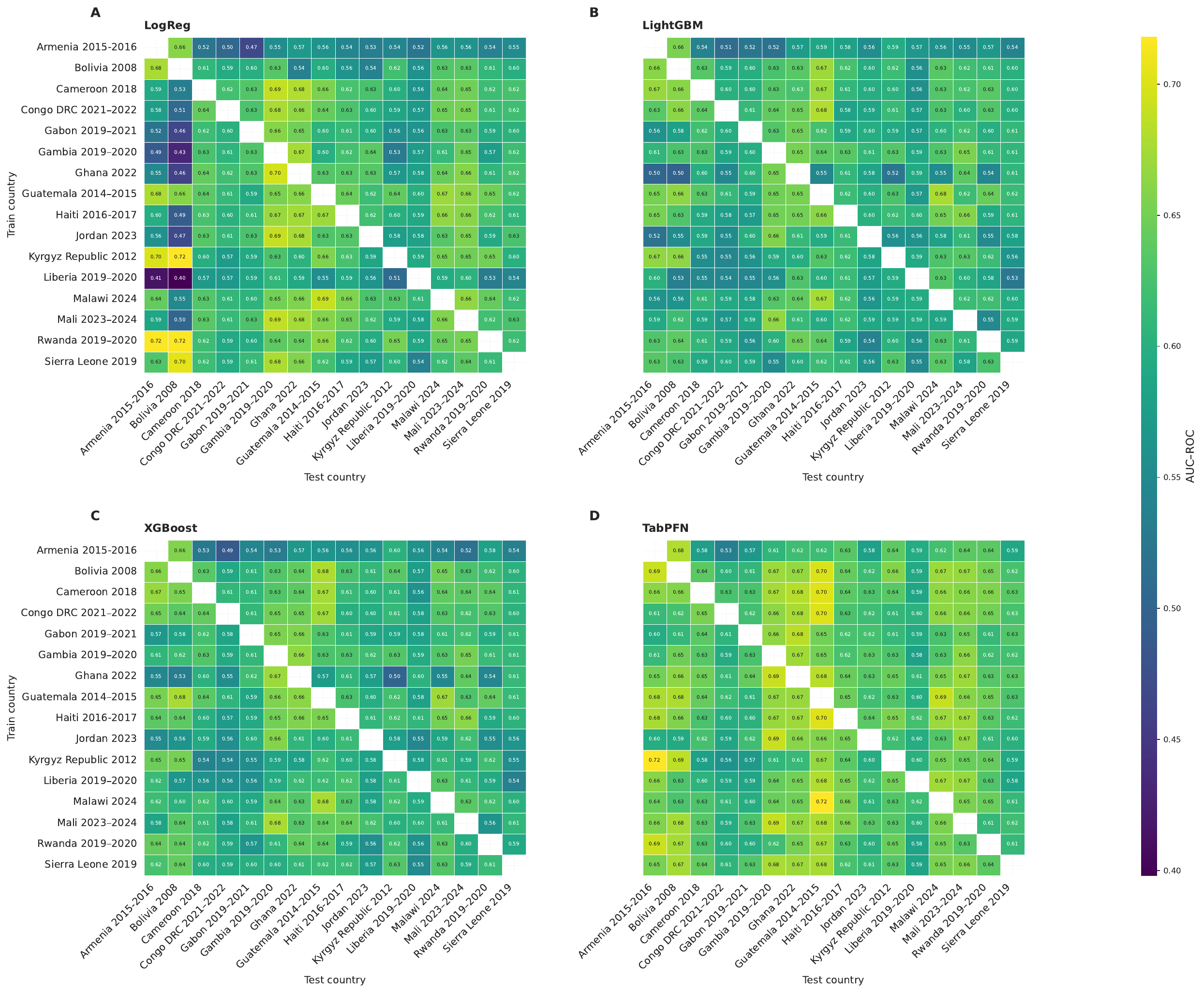}
\caption{\textbf{Cross-country transferability of predictive models under reverse LOCO validation.} Each panel shows AUC-ROC for a given model (A: Logistic Regression; B: LightGBM; C: XGBoost; D: TabPFN~v2.6) across all 240 directed train--test country pairs. Rows denote the training country; columns denote the held-out test country; diagonal cells are blank (same-country pairs excluded). Color intensity reflects AUC-ROC on a shared scale (dark blue $\approx$0.40; yellow $\ge$0.70). Row-wise and column-wise variability indicate asymmetric transferability: some countries generalize better as training sources, while others are consistently harder to predict as target populations, independent of the model used.}
\label{fig:reverse_loco_plot}
\end{figure}

Training on a single source country and evaluating on each of the remaining 15 revealed substantial regularities in cross-country transferability across all four models (Figure~\ref{fig:reverse_loco_plot}). AUC-ROC values under this setting were generally lower and more variable than under the standard LOCO framework, as expected given that models were trained on a single source population rather than a pooled multi-country setup. Most train--test country pairs produced AUC-ROC values in the range of 0.55--0.72. Performance varied considerably by both source and target country, and this variation exceeded inter-model differences throughout. Countries with larger and epidemiologically diverse populations (e.g.\ Ghana, Guatemala, Haiti, Bolivia) served as more informative source domains, producing comparatively higher AUC-ROC values across multiple target countries. By contrast, models trained on smaller or more epidemiologically atypical populations (e.g.\ Armenia, Jordan) transferred poorly to most other settings, with some target-country AUC-ROC values as low as 0.40 for Logistic Regression. Directionality was evident throughout: transfer from country A to country B did not imply equivalent transfer in the reverse direction, indicating that predictive signal is not symmetrically shared between populations. Column-wise patterns, in which certain countries were consistently difficult to predict regardless of the training source, further suggest that target-side characteristics such as local prevalence, predictor distributions, and data quality impose an independent ceiling on transferability. This structural pattern was broadly consistent across all four models, though TabPFN showed somewhat more heterogeneity in its off-diagonal values, including occasional instances of higher transfer in specific country pairs.

\subsection{Subgroup performance analysis}
\label{subsec:subgroup}

\begin{figure}[!ht]
\centering
\includegraphics[width=\textwidth]{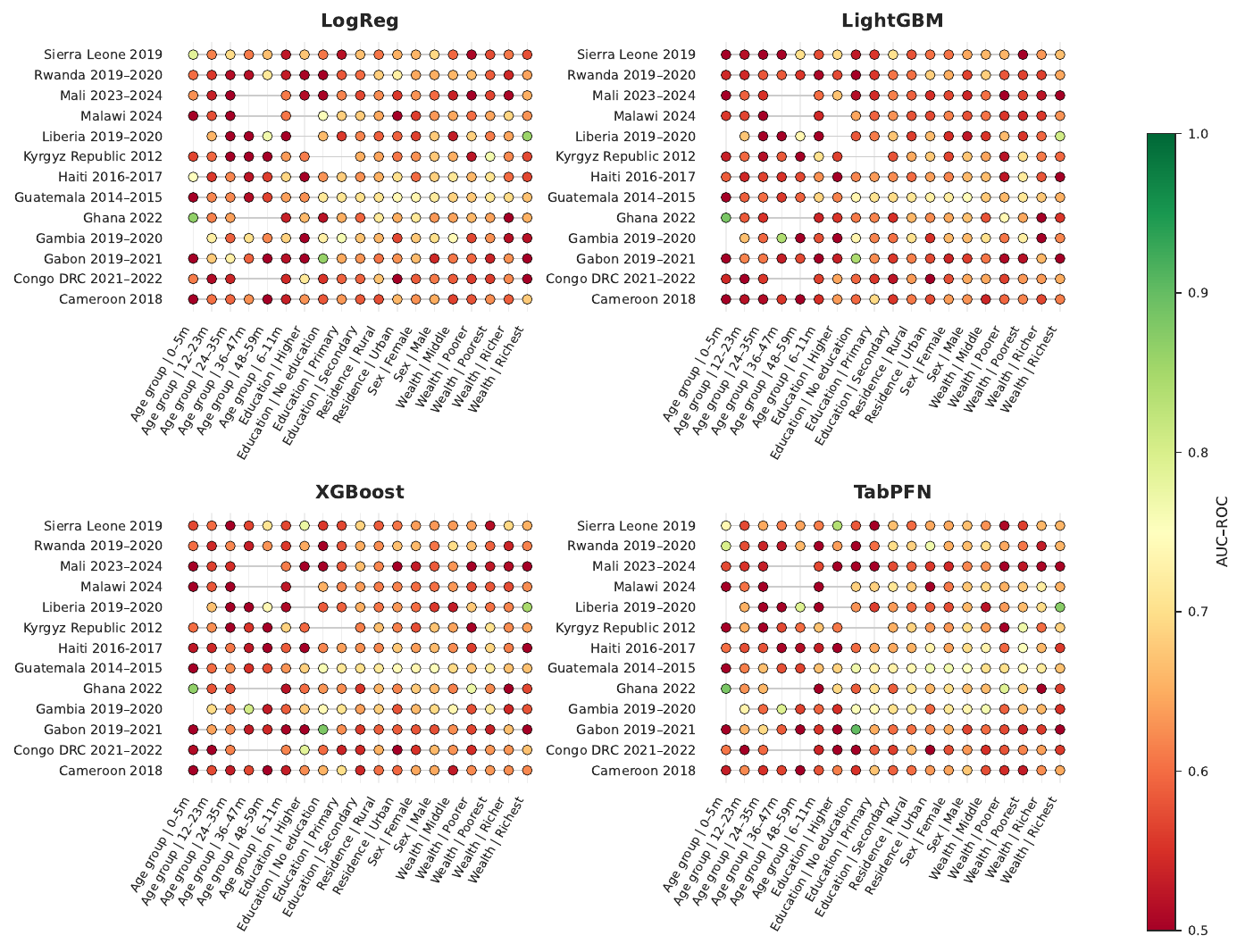}
\caption{\textbf{Subgroup discriminative performance (AUC-ROC) across demographic strata, countries, and models.} Each panel shows AUC-ROC for a given model across available demographic subgroups (columns) and countries (rows), evaluated under leave-one-country-out validation. Subgroups include child age group in months, maternal education level (no education, primary, secondary, higher), residence type (rural, urban), child sex (female, male), and household wealth quintile (poorest to richest). Colour encodes AUC-ROC on a shared scale (red $\approx$0.50; green $\approx$1.00). Empty cells indicate strata excluded from analysis due to absence of both outcome classes after filtering. Armenia, Bolivia, and Jordan are excluded entirely as no ML-valid subgroup partitions were available after preprocessing.}
\label{fig:fig_subgroup_dotline_plot}
\end{figure}

Model performance was evaluated across five demographic strata: child age group, maternal education level, residence type, child sex, and household wealth quintile (Figure~\ref{fig:fig_subgroup_dotline_plot}). Subgroup analyses were not feasible in Armenia, Bolivia, and Jordan due to insufficient sample sizes within one or more strata after outcome exclusion and age restriction (Appendix~\ref{tab:subgroup_counts}), reflecting data sparsity in those settings rather than systematic model behavior. Across the remaining 13 countries, AUC-ROC values were broadly similar across demographic subgroups within the same country, with no model exhibiting consistent or pronounced bias toward specific strata. The dominant source of variation was again country and subgroup characteristics rather than model choice. Greater within-country variability was observed across age, maternal education, and household wealth strata. Countries including Sierra Leone, Liberia, Ghana, Gabon, and The Gambia showed the largest subgroup differences, most notably between younger (0--5 months) and older (36--47 months) age groups. In these settings, the absence of maternal formal education was associated with more pronounced differences in predictive performance, and the highest wealth quintile in Liberia showed comparatively higher discrimination across all models. Residence-based subgroups showed limited systematic differences, and wealth quintiles were broadly uniform across most other countries with no consistent gradient from poorest to richest. Inter-model differences within each country--subgroup combination remained small and within a narrow AUC range, mirroring patterns observed in the overall and LOCO analyses.

\subsection{Decision curve analysis}
\label{subsec:dca}
\begin{figure}[!ht]
\centering
\includegraphics[width=0.75\textwidth]{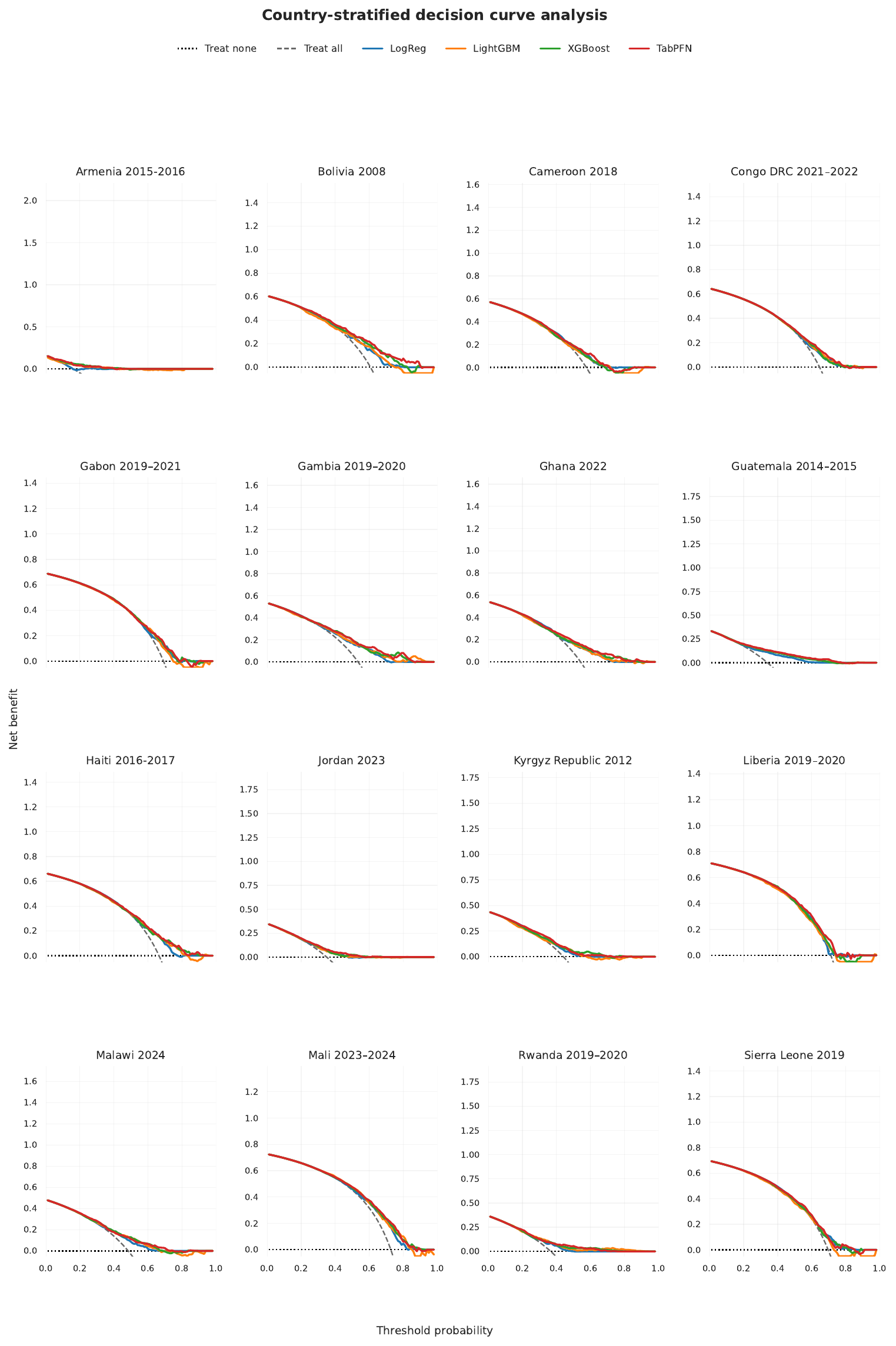}
\caption{\textbf{Country-stratified decision curve analysis for four predictive models of childhood Anemia.} Net benefit is shown as a function of threshold probability for all models alongside reference strategies of treating all children (dashed grey) and treating none (dotted black). Each panel represents one of the 16 study countries, ordered by continental group. All models outperform the treat-none strategy across a wide range of threshold probabilities, and all outperform treat-all above low thresholds. Inter-model differences in net benefit are narrow throughout the clinically plausible range; country-level variation in curve shape and magnitude reflects differences in Anemia prevalence across settings. Net benefit values below zero are truncated at the axis for clarity.}
\label{fig:figure_dca_country_clean}
\end{figure}

Decision curve analysis assessed the clinical utility of each model across the range of threshold probabilities relevant to Anemia screening decisions. Across all 16 countries, all four models provided positive net benefit over the treat-none strategy across a broad range of threshold probabilities, and all outperformed the treat-all strategy above low thresholds (Figure~\ref{fig:figure_dca_country_clean}). Inter-model differences in net benefit were narrow throughout the clinically plausible range, with the four model curves overlapping substantially and no single model demonstrating a consistent advantage. This pattern mirrors the discrimination and calibration analyses: country-level variation exceeded inter-model variation as the dominant source of heterogeneity, and the threshold at which action is triggered and the population to which the model is applied are stronger determinants of net benefit than model choice. In high-prevalence settings such as Congo DRC, Mali, and Gambia, net benefit remained positive over a wide range of threshold probabilities, reflecting the greater opportunity for risk-stratified intervention when the prior probability of Anemia is high. In lower-prevalence settings such as Jordan and Armenia, net benefit curves declined more steeply and converged toward zero at lower thresholds, consistent with a narrower decision-relevant window.

\subsection{Interpretability analysis}
\label{subsec:interpretability}

\begin{figure}[!ht]
\centering
\includegraphics[width=1.0\textwidth]{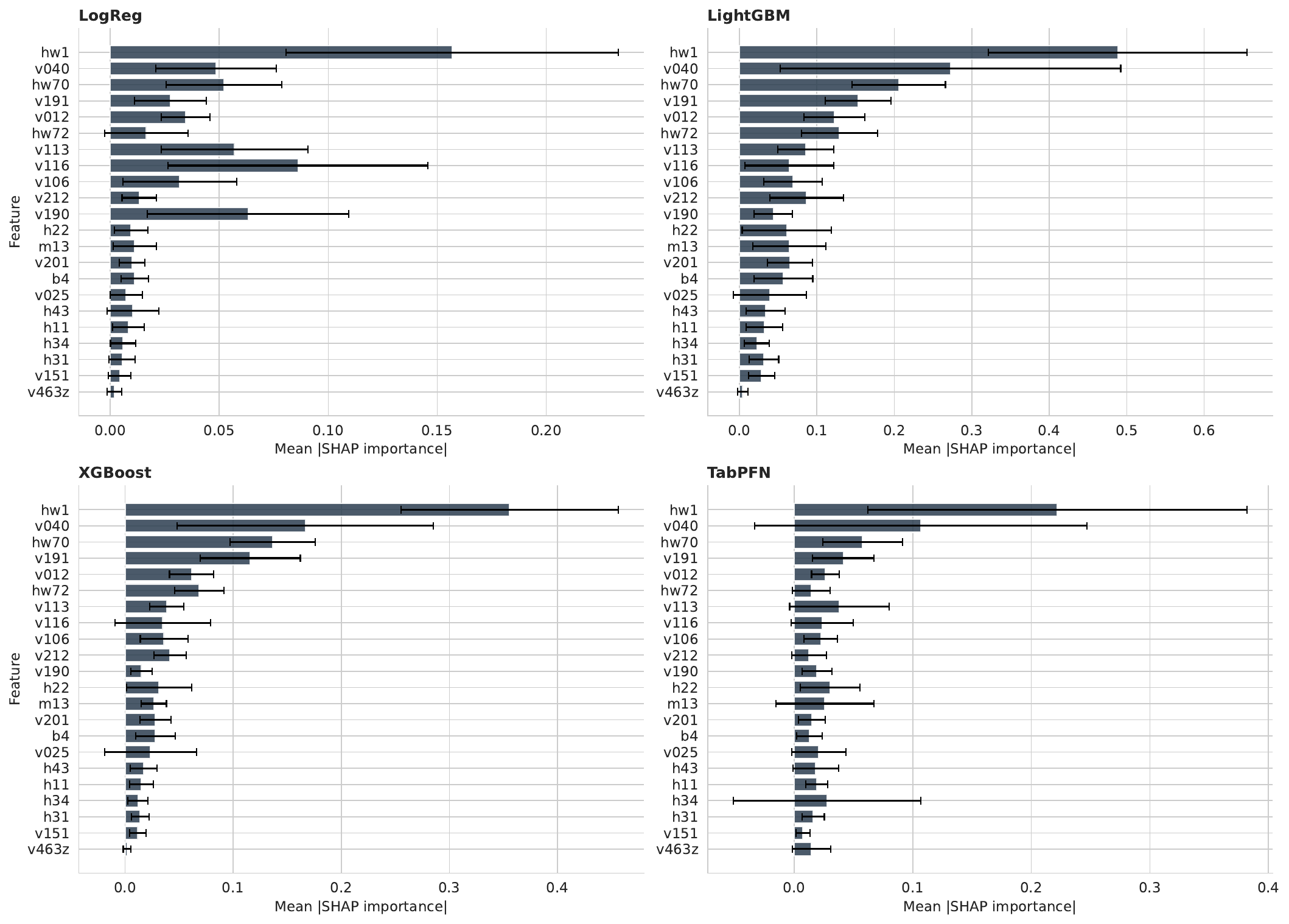}
\caption{\textbf{Aggregated feature importance across the four predictive models.} Mean SHAP importance values (with standard deviation) are shown for all models, averaged across all 16 study countries. Features are ordered by descending mean importance within each model panel. Variable codes correspond to DHS predictors described in Table~\ref{tab:features}: \texttt{hw1} = child age; \texttt{v040} = altitude; \texttt{hw70} = height-for-age z-score; \texttt{v191} = wealth index; \texttt{v012} = maternal age; \texttt{v106} = maternal education; \texttt{hw72} = weight-for-age z-score. Error bars indicate variability in feature importance scores.}
\label{fig:feature_importance_plot}
\end{figure}

\begin{figure}[!ht]
\centering
\includegraphics[width=0.88\textwidth]{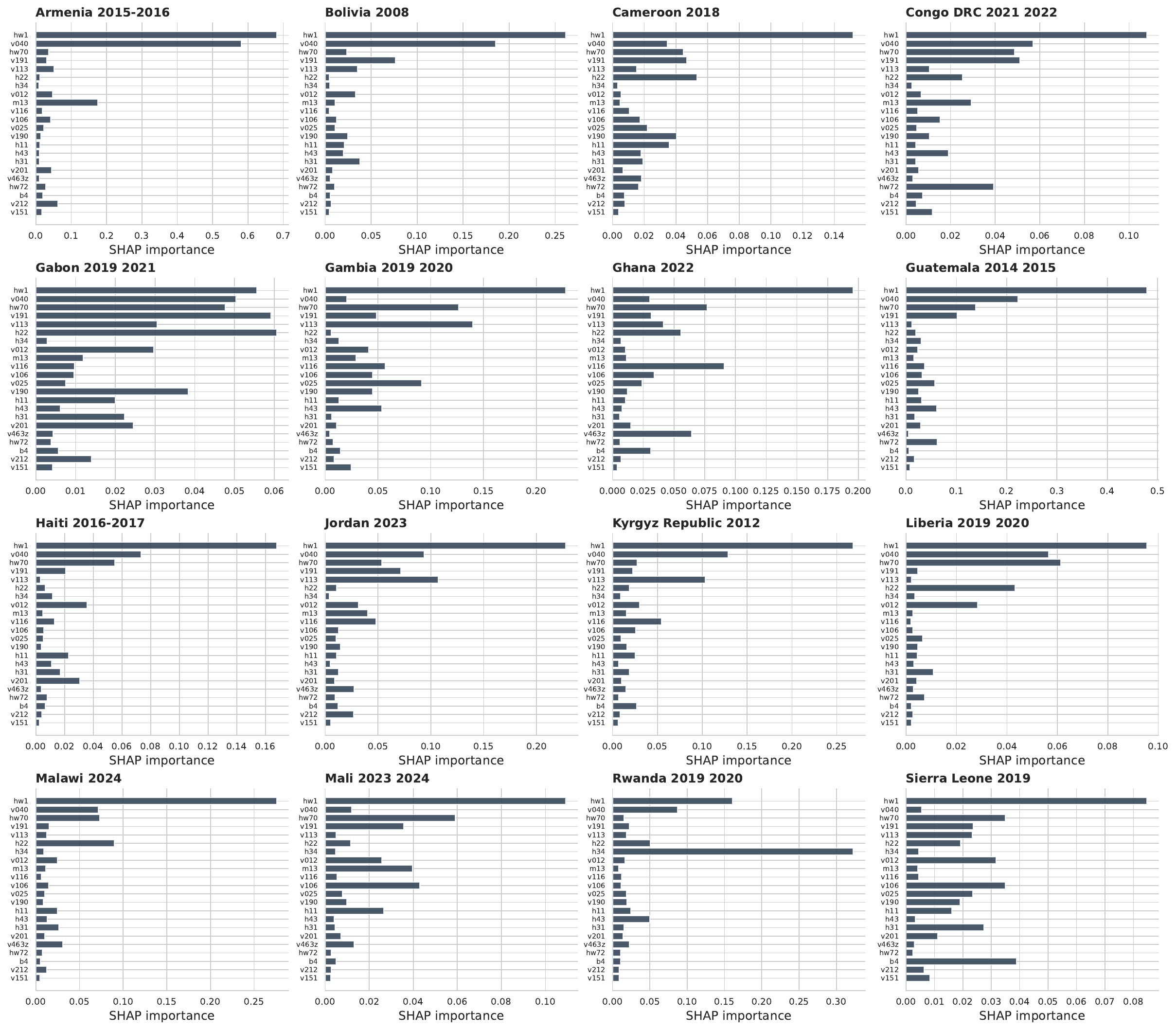}
\caption{\textbf{Country-level feature importance for TabPFN~v2.6 across 16 study populations.} Each panel shows  importance for each predictor within a single country, estimated by measuring the decline in AUC-ROC when each feature column is randomly permuted on the held-out test set under within-country evaluation. Features are ordered consistently across panels to facilitate comparison. Child age (\texttt{hw1}) was the dominant predictor in most settings; the relative contribution of altitude (\texttt{v040}), anthropometric z-scores (\texttt{hw70}, \texttt{hw72}), and socioeconomic variables (\texttt{v191}, \texttt{v106}) varied across countries, reflecting heterogeneity in the epidemiological drivers of Anemia across the included populations.}
\label{fig:tabpfn_feature_importance_countries}
\end{figure}

Feature importance analyses examined which predictors drove model decisions across countries and methods. For all four models, SHAP values were computed and aggregated across all countries: TreeSHAP was used for LightGBM and XGBoost, standard linear SHAP for logistic regression, and the shapiq remove-and-recontextualize framework for TabPFN, which adapts Shapley-based attribution to account for the in-context inference mechanism underlying TabPFN predictions. Results are summarized in the aggregated plot (Figure~\ref{fig:feature_importance_plot}) and the country-specific TabPFN analysis (Figure~\ref{fig:tabpfn_feature_importance_countries}), with additional model-level results provided in~\ref{app:other_model_interpretability}. Child age (\texttt{hw1}) was the dominant predictor across all four models, with the largest mean SHAP importance in every model panel. Altitude (\texttt{v040}) ranked consistently second across logistic regression, LightGBM, and XGBoost, and remained among the top features for TabPFN, reflecting the role of altitude-adjusted hemoglobin thresholds in the outcome definition. Height-for-age z-score (\texttt{hw70}), wealth index (\texttt{v191}), and maternal education (\texttt{v106}) appeared in the upper tier of importance across all models, consistent with the known epidemiology of childhood Anemia. Weight-for-age (\texttt{hw72}) and weight-for-height z-scores contributed meaningfully but with greater variability across models. Features related to illness (\texttt{h11}, \texttt{h31}), deworming (\texttt{h43}), vitamin A supplementation (\texttt{h34}), and smoking (\texttt{v463z}) were consistently ranked in the lower tier, suggesting modest marginal contributions after conditioning on age, anthropometry, and socioeconomic variables. The rank ordering of features was broadly stable across all four models, with TabPFN showing a qualitatively similar ordering to the supervised models, providing cross-method convergent validity for the identified predictor hierarchy. Country-level TabPFN SHAP profiles revealed meaningful heterogeneity in the relative contribution of individual features: child age was particularly pronounced in Guatemala, Rwanda, and Liberia, while in Armenia and Bolivia the importance distribution was more diffuse, consistent with smaller and more homogeneous samples in those settings. Altitude (\texttt{v040}) showed high SHAP importance in settings with substantial within-country elevational variation, including Bolivia and Guatemala (Andean and highland regions), the Kyrgyz Republic (Central Asian mountain terrain), Rwanda (high-plateau geography), and Jordan, where altitude-adjusted hemoglobin thresholds produce meaningful variation in the derived outcome. By contrast, altitude contributed little in coastal or uniformly low-elevation settings such as Gambia, Sierra Leone, and Liberia, where elevational range is narrow and the altitude adjustment introduces minimal outcome heterogeneity. This pattern reflects the structural role of altitude in the outcome definition rather than a direct causal pathway, and its cross-country variability in importance is therefore an expected consequence of geography rather than a signal of differential biological relevance. These patterns are broadly consistent with the known epidemiological literature. These analyses are descriptive and do not support causal inference; feature importance reflects statistical association with the outcome conditional on other predictors in the training data and should not be interpreted as evidence of causal pathways.

\section{Discussion}
\label{sec:discussion}

\subsection{Foundational models in data-scarce and low-resource settings}

The most consistent and practically meaningful difference between models emerged not under full-data conditions, but in the low-data regime. Below circa 200 within-country training samples, TabPFN consistently outperformed Logistic Regression, XGBoost, and LightGBM, with the advantage most pronounced in Armenia and Liberia where small sample sizes and lower Anemia prevalence place the hardest constraints on conventional supervised learning. This finding is theoretically coherent: rather than learning from scratch on a fixed training set, TabPFN arrives at inference with representations shaped by pre-training on a large and diverse corpus of synthetic tabular tasks, encoding generic statistical relationships across a wide range of feature structures and outcome types~\cite{hollmann2025tabpfn}. In the context of childhood Anemia prediction, where labeled data are often sparse in the settings where surveillance tools are most urgently needed, this background knowledge functions as a form of implicit regularization, stabilizing inference when local data are insufficient to reliably identify predictive structure. As labeled data accumulated, the gap over well-tuned gradient-boosted models narrowed considerably, a theoretically expected property of any prior-regularized estimator rather than a limitation of the approach.

TabPFN's advantage in probabilistic calibration is equally consequential for public health practice. Across all 16 countries, TabPFN achieved the lowest mean Brier score (0.203) and mean ECE (0.042), outperforming XGBoost (0.207; 0.048), Logistic Regression (0.212; 0.052), and LightGBM (0.213; 0.069). A well-calibrated model ensures that a predicted probability of Anemia corresponds more faithfully to the observed frequency of anemic children in that risk stratum, a property that matters directly when model outputs inform screening thresholds or resource allocation decisions in settings with limited diagnostic infrastructure~\cite{riley2016external}. LightGBM exhibited the most visible miscalibration across countries, particularly at higher predicted probabilities, while TabPFN produced curves closest to the diagonal across a range of prevalence settings. That said, TabPFN's reduced transparency relative to logistic regression and its dependence on external model infrastructure may constrain adoption in settings where interpretability is a regulatory or community trust requirement, and where data residency and national ownership of clinical decision tools are increasingly mandated. These practical constraints represent important considerations for deployment and warrant dedicated attention in future implementation work.

\subsection{Generalization performance and distributional shift}

Beyond the low-data regime, within-country discrimination was moderate across all four models, with AUC-ROC values spanning 0.59 to 0.76, and the gap between the best and worst model within any single country rarely exceeding 0.05 AUC points. This reflects the inherent difficulty of predicting a multi-factorial outcome from a standardized but necessarily coarse set of survey-derived predictors. Anemia in young children arises from a confluence of nutritional deficiencies, infectious burden, maternal and household factors, and environmental exposures whose relative contributions vary substantially across ecological zones, healthcare systems, and economic contexts~\cite{who_anemia_2025,gardner2023prevalence}. The highest within-country discrimination occurred in Armenia and Guatemala, while Liberia and Sierra Leone remained persistently difficult to predict across all models, consistent with smaller effective sample sizes, higher predictor missingness, and Anemia etiologies partly driven by endemic malaria and hemoglobinopathies that are not directly captured by available DHS predictors.

A critical but often underappreciated source of this heterogeneity is the data-generating process itself. DHS surveys in our cohorts span 2008 to 2024, and both survey methodology and the epidemiological profiles of target populations evolve over time. Temporal shifts in Anemia burden driven by changes in malaria control programs, food fortification policies, or healthcare access are not captured by any static cross-sectional model. Beyond temporal factors, the quality of DHS data varies across settings in ways that are difficult to quantify but are epidemiologically consequential: field worker competence, hemoglobin measurement protocols, and biomarker reporting fidelity may differ across survey rounds and implementing organizations, directly affecting the predictive signal available to any model. The structure of the analytic cohort compounds this further. Children with available hemoglobin measurements are not a simple random sample of all children in those settings, as selection into biomarker measurement may be associated with healthcare access, household socioeconomic position, or field logistics, introducing country-varying non-representativeness. Over half of the retained records (53.3\%) required imputation for at least one predictor, and patterns of missingness themselves reflect structural differences in survey implementation. In settings with lower human capital indices and higher rates of illiteracy, self-reported variables such as illness history or vitamin supplementation may carry greater measurement error, attenuating the informational value of predictors that perform well elsewhere. These considerations collectively caution against assuming that a model trained on one or several DHS surveys can be straightforwardly deployed in a new setting, consistent with the broader literature on shortcut learning and external validity in clinical prediction models~\cite{riley2016external,geirhos2020shortcut,ong2024shortcut}.

Under LOCO validation, performance declined modestly but uniformly across all models, with held-out AUC-ROC values clustering between 0.62 and 0.66 for most countries. Guatemala consistently yielded the highest external AUC-ROC (0.68--0.69), plausibly reflecting its large analytic sample, relatively complete biomarker coverage, and nutritional and socioeconomic risk profile proximate to the pooled training distribution. Performance was lowest when Ghana or Liberia was held out, consistent with the within-country findings. Inter-model differences under LOCO remained small and inconsistent, reinforcing that target-population characteristics rather than model architecture were the primary determinant of cross-country generalization.

\subsection{Knowledge transfer and asymmetry}

The reverse LOCO analyses revealed a dimension of cross-population transferability that aggregated LOCO statistics obscure: directionality matters. Training on a single source country and evaluating on each of the remaining 15 produced AUC-ROC values ranging from 0.55 to 0.72, with variation driven far more by the identity of the source and target countries than by model choice. Larger and epidemiologically diverse populations, including Ghana, Guatemala, Haiti, Cameroon, Congo DRC, and Bolivia, served as more informative source domains, producing comparatively higher AUC-ROC values across multiple target countries. By contrast, models trained on Armenia transferred poorly to most other settings, with some AUC-ROC values approaching 0.50 for logistic regression. This asymmetry suggests that the predictive signal embedded in a country's data reflects its epidemiological breadth: populations with heterogeneous Anemia determinants spanning nutritional, infectious, and socioeconomic axes appear to produce representations that generalize more broadly than those from smaller, lower-prevalence, or more homogeneous settings. Column-wise patterns in the reverse LOCO matrices, where certain countries were consistently difficult to predict regardless of the training source, further suggest that target-side characteristics including local prevalence, predictor distributions, and data quality impose an independent ceiling on transferability. Where Anemia is driven by highly context-specific factors such as endemic malaria or locally dominant dietary deficiencies, cross-country training data from dissimilar populations may not recover that signal regardless of sample size or model complexity. These findings caution against naive data-pooling strategies: simply aggregating DHS surveys from multiple countries does not guarantee equitable generalization, as the effective influence of each source country on model behavior is unequal, transfer is asymmetric, and temporal distance between survey rounds introduces additional drift. From a global health equity perspective, the asymmetry we document highlights a structural risk: the countries with the sparsest surveillance data and the most limited biomarker coverage are precisely those for which externally trained models are least reliable, and for which the few-shot advantage of TabPFN documented here is most operationally relevant.

\subsection{Limitations and future directions}

This study has some limitations. First, all analyses relied on DHS data, limiting predictions to current Anemia status and precluding assessment of temporal dynamics or causal pathways. Second, despite spanning 16 countries across five regions, the sample is constrained to settings with available DHS surveys and may not fully represent broader global heterogeneity, including conflict-affected or high-income contexts. Third, the harmonized predictor set, while necessary for cross-country comparability, excludes potentially important context-specific covariates such as malaria endemicity, dietary intake measures, or local hemoglobinopathy prevalence that likely drive the residual performance variability we observe. Fourth, while TabPFN demonstrated meaningful advantages in low-data regimes and calibration, its long-term maintainability in operational public health settings warrants scrutiny: adapting foundation models to local epidemiological shifts requires access to model internals and substantial computational infrastructure, creating dependencies on external providers that may conflict with growing requirements for AI sovereignty, data residency, and national auditability of clinical decision tools. Finally, our modeling framework assumes independent and identically distributed data within each training set, an assumption that does not fully hold given the distributional shifts documented across populations and survey rounds. Several directions could extend this work. Incorporating geospatial, climatic, or contextual data modalities may improve robustness by capturing environmental drivers of Anemia not available in standard DHS instruments. Domain adaptation and transfer learning frameworks offer a principled path toward addressing the cross-population distribution shifts identified here, and treating source-country selection and weighting as explicit modeling decisions rather than uniform pooling deserves formal investigation. Advances in self-supervised and multimodal foundation models for health data may enable more transferable representations without reliance on labeled data. Finally, causal representation learning approaches may help move beyond statistical associations toward modeling underlying mechanisms, improving both robustness and interpretability across diverse settings. Realizing the practical value of predictive tools for childhood Anemia screening will ultimately require not only methodological advances but sustained investment in local data infrastructure, capacity building, and context-sensitive validation pipelines.

\section{Conclusion}
\label{sec:conclusion}
In this extensive multi-country study spanning 16 nationally representative DHS cohorts across diverse global settings, we systematically investigated a broad range of machine learning approaches, including conventional learning methods and emerging transformer-based tabular foundation models for childhood Anemia prediction. Beyond predictive discriminability, the study examined model robustness and generalization across populations, calibration behavior, subgroup performance across demographic and clinical strata, feature-level interpretability, and potential clinical utility through DCA.

Collectively, our findings demonstrate that predictive performance for childhood Anemia remains strongly influenced by population heterogeneity and cross-country distributional differences, with no single modeling approach achieving uniformly optimal performance across all data-rich scenarios. Several factors appeared to constrain generalization, including heterogeneous Anemia etiology across populations, asymmetric transferability between countries, temporal shifts, variable data quality, and the limited representativeness inherent to survey-based datasets. Importantly, these constraints are unlikely to be resolved through algorithmic innovation alone, underscoring the need for sustained investment in local data infrastructure, interdisciplinary collaboration, context-sensitive validation frameworks, multimodal integration, and transfer learning strategies that explicitly treat population heterogeneity as a central modeling consideration.

Nevertheless, tabular foundation models consistently showed competitive discrimination, improved calibration characteristics, and greater stability under data-limited regimes, suggesting that these models, owing in part to the broad background knowledge acquired during pre-training, may offer important vistas of opportunity in heterogeneous and resource-constrained environments where reliable probability estimation is critical for downstream decision-making. The subgroup and interpretability analyses further indicated that model behavior was broadly aligned with established epidemiological determinants of childhood Anemia, while also highlighting persistent variability in performance across population strata. Together, these findings reinforce the necessity of evaluating transportability, calibration, and subgroup robustness alongside conventional performance metrics when developing predictive systems intended for utilization across diverse global health settings.

Overall, our study supports the growing potential of transformer-based tabular foundation models as clinically relevant tools for risk stratification and screening applications in digital health. Within the practical constraints of low-resource environments, these models represent a promising and calibration-sound pathway for Anemia prediction, particularly in settings where robust estimation under limited supervision is paramount.

\section*{Data and Code Availability}
The data used in this study are available from the Demographic and Health Surveys (DHS) Program (\url{https://dhsprogram.com/Data/}) upon registration and request. Due to data use agreements, the datasets cannot be redistributed by the authors.

All code for data processing, model development, and analysis is publicly available at:
\url{https://github.com/yusufbrima/anemiafoundationmodel}

\section*{Author Contributions (CRediT Taxonomy)}
\noindent \textbf{Conceptualization:} YB \\
\noindent \textbf{Methodology:} YB \\
\noindent\textbf{Formal Analysis:} YB, MA \\
\noindent\textbf{Investigation:} YB, MA, LHK\\
\noindent\textbf{Data Curation:} YB \\
\noindent\textbf{Coding and Visualization:} YB\\
\noindent\textbf{Writing - Original Draft:} YB, MA, LHK, DN, AV, SS, DGC \\
\noindent\textbf{Writing - Review \& Editing:} YB, MA, LHK, DN, AV, SS, DGC \\
\noindent\textbf{Supervision:} MA, DGC \\
\noindent\textbf{Project Administration:} YB, MA \\
\noindent\textbf{Funding Acquisition:} MA, DGC

\section*{Declaration of interests}
\noindent We declare no competing interests.

\section*{Ethics statement}
This study uses anonymized secondary data from the DHS Program. Ethics approval for the original surveys was obtained by the DHS Program. Secondary analysis of de-identified data will be handled according to institutional requirements.


\bibliographystyle{vancouver}
\bibliography{references}

\clearpage
\appendix

\section{Preprocessing, modelling and evaluation}
\subsection{Preprocessing pipeline}
\label{app:cohort_derivation_flowchart}
\begin{figure}[H]
\centering
\includegraphics[width=\textwidth]{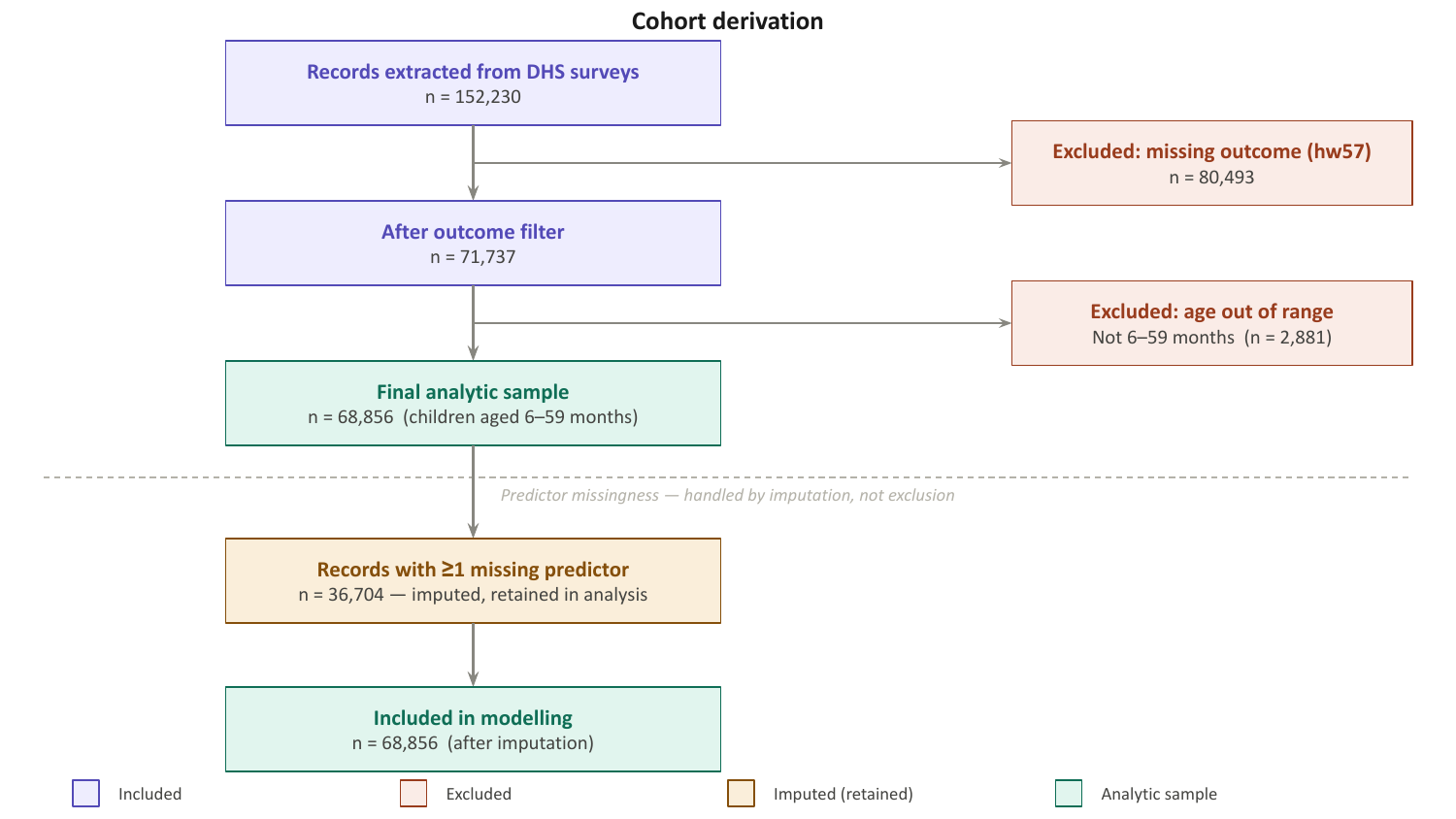}
\caption{\textbf{Cohort derivation flowchart.} Records were drawn from DHS across all included countries (\textit{n}\,=\,152{,}230). Records with missing Anemia status (hw57) were excluded (\textit{n}\,=\,80{,}493), followed by exclusion of children outside the 6--59\,month age eligibility window (\textit{n}\,=\,2{,}881), yielding a final analytic cohort of 68{,}856 children. Of these, 36{,}704 records (53.3\%) had at least one missing predictor value; missing data were handled by median imputation for continuous variables and mode imputation for categorical variables prior to model training.}
\label{fig:cohort_derivation_flowchart}
\end{figure}

DHS child recode (KR) datasets were harmonized using standardized variable mappings. DHS missing codes were recoded as missing. Age (hw1) was used to restrict the sample  to children aged 6--59 months, and observations with missing outcome data (hw57) were excluded, yielding a final analytic sample of 68,856 children (Figure~\ref{fig:cohort_derivation_flowchart}; Table~\ref{tab:exclusion_counts}). Categorical variables were encoded numerically within the modelling pipeline. Continuous variables were standardized using z-scores based on training data. Missing predictor values, present in 36,704 records (53.3\%), were imputed using median (continuous) and mode (categorical) imputation prior to model training.

\begin{table}[ht]
\centering
\caption{Cohort derivation across DHS countries}
\label{tab:exclusion_counts}
\begin{tabular}{lrrrrrrr}
\toprule
Country & Raw N & After Outcome & Excl. Outcome & After Age & Excl. Age & Final N & Missing Predictors \\
\midrule

Gabon 2019--2021        & 6,376  & 4,659  & 1,717  & 4,462  & 197  & 4,462  & 2,017 \\
Malawi 2024             & 12,312 & 4,081  & 8,231  & 3,923  & 158  & 3,923  & 2,125 \\
Rwanda 2019--2020       & 8,092  & 3,436  & 4,656  & 3,303  & 133  & 3,303  & 1,032 \\
Sierra Leone 2019       & 9,899  & 3,510  & 6,389  & 3,367  & 143  & 3,367  & 1,142 \\
Haiti 2016--2017        & 6,530  & 4,934  & 1,596  & 4,773  & 161  & 4,773  & 1,786 \\
Jordan 2023             & 9,106  & 4,373  & 4,733  & 4,209  & 164  & 4,209  & 4,209 \\
Armenia 2015--2016      & 1,724  & 1,361  & 363    & 1,313  & 48   & 1,313  & 1,313 \\
Kyrgyz Republic 2012    & 4,363  & 3,617  & 746    & 3,437  & 180  & 3,437  & 1,268 \\
Liberia 2019--2020      & 5,704  & 2,154  & 3,550  & 2,048  & 106  & 2,048  & 737 \\
Bolivia 2008            & 8,605  & 2,378  & 6,227  & 2,233  & 145  & 2,233  & 2,233 \\
Congo DRC 2021--2022    & 24,000 & 9,365  & 14,635 & 9,002  & 363  & 9,002  & 6,207 \\
Guatemala 2014--2015    & 12,440 & 10,385 & 2,055  & 9,987  & 398  & 9,987  & 3,378 \\
Ghana 2022              & 9,353  & 3,882  & 5,471  & 3,742  & 140  & 3,742  & 1,948 \\
Gambia 2019--2020       & 8,362  & 3,274  & 5,088  & 3,146  & 128  & 3,146  & 1,376 \\
Cameroon 2018           & 9,733  & 3,987  & 5,746  & 3,825  & 162  & 3,825  & 1,922 \\
Mali 2023--2024         & 15,631 & 6,341  & 9,290  & 6,086  & 255  & 6,086  & 4,011 \\
\bottomrule
\end{tabular}
\end{table}

\subsection{Computational environment and runtime}
All analyses were implemented in Python 3.10 using scikit-learn, LightGBM, XGBoost, and Optuna, and executed on a single NVIDIA A100 (80GB) GPU system. Runtime varied by model, with logistic regression completing in under one minute, LightGBM and XGBoost requiring approximately 5-20 minutes depending on hyperparameter settings, and TabPFN performing inference in a few seconds per batch due to its in-context learning design.

\subsection{Hyperparameter optimization (Optuna)}
\label{app:hyperparameter_search}
We used Bayesian (Tree-structured Parzen Estimator) optimization with 30 trials per model to tune hyperparameters. Search spaces were:

\begin{longtable}{p{0.22\textwidth}p{0.72\textwidth}}
\toprule
\textbf{Model} & \textbf{Hyperparameters (range or options)} \\
\midrule
Logistic regression & Regularization strength $C \sim \text{logU}[10^{-3},10]$; solver $\in \{\text{liblinear}, \text{lbfgs}\}$; class\_weight $\in \{\text{none}, \text{balanced}\}$. \\
LightGBM & n\_estimators: [100, 600]; learning\_rate: logU($10^{-2},0.2$); num\_leaves: [16,128]; max\_depth: $\{-1,3,5,7,10\}$; subsample: [0.6,1.0]; colsample\_bytree: [0.6,1.0]. \\
XGBoost & n\_estimators: [100, 600]; learning\_rate: logU($10^{-2},0.2$); max\_depth: [3,10]; subsample: [0.6,1.0]; colsample\_bytree: [0.6,1.0]. \\
\bottomrule
\end{longtable}

\noindent Logistic regression had no hyperparameters other than regularization.

\subsection{Few-shot learning evaluation}
We simulated low-resource data availability by limiting the fraction of country-specific labels for training. Five regimes were considered: 1\%, 5\%, 10\%, and 20\% target-country samples. In each scenario, the target country's data was randomly split into a small training set (the specified percentage) and a held-out test set. Models were trained on the small training set and evaluated on the remaining held-out target data. TabPFN was given the (tiny) training set as its in-context learning. We computed AUC, F1, sensitivity, specificity, and accuracy for each regime. This analysis reflects realistic resource-constrained settings where only a small labeled sample is available in a new country. (See Appendix Table~\ref{tab:subgroup_counts} for detailed results.) \\
\emph{Limitations:} Performance estimates at very low shot levels (e.g.\ 1\%) have higher variance due to the tiny training set; results should be interpreted accordingly.

\subsection{Leave-One-Country-Out (LOCO) framework}
Let $\mathcal{D}_1, \ldots, \mathcal{D}_{16}$ denote the country-specific datasets. In the LOCO evaluation, for each $k=1,\ldots,16$, we define the training set as $\mathcal{D}_{-k} = \bigcup_{j\neq k}\mathcal{D}_j$ (all countries except $k$) and the test set as $\mathcal{D}_k$ (the held-out country). Models are trained on $\mathcal{D}_{-k}$ and evaluated on $\mathcal{D}_k$, iteratively for each country. This procedure tests how well a model generalizes to a completely unseen population.

\subsection{Additional performance metrics for predictive evaluation}
In addition to AUROC, we computed:
\begin{itemize}
  \item Accuracy, precision, recall, and F1-score at the thresholds.
\end{itemize}
These metrics were summarized per country and averaged ($\pm$SD) across countries to supplement the main results.

\subsection{Calibration metrics and mathematical definitions}
We assessed calibration with the \textbf{Brier score} and \textbf{Expected Calibration Error (ECE)}. For a dataset of size $N$ with true outcomes $y_i\in\{0,1\}$ (where $y_i=1$ denotes Anemia and $y_i=0$ denotes no Anemia) and corresponding model-predicted probabilities $\hat{p}_i\in[0,1]$ for child $i$, the Brier score is defined as:
\[
\text{Brier} = \frac{1}{N}\sum_{i=1}^{N}(y_i-\hat{p}_i)^2.
\]
This is the mean squared error between predicted probabilities and observed binary outcomes; a perfectly calibrated model with no uncertainty would achieve a Brier score of 0.

ECE is computed by partitioning all $N$ predictions into $B$ equal-width bins over the interval $[0,1]$. For each bin $b\in\{1,\ldots,B\}$ containing $n_b$ observations, $\text{acc}(b)$ denotes the empirical proportion of Anemia cases (i.e.\ the observed frequency of $y=1$) and $\text{conf}(b)$ denotes the mean predicted probability $\hat{p}$ within that bin. ECE is then the sample-size-weighted average of the absolute difference between confidence and accuracy across bins:
\[
\text{ECE} = \sum_{b=1}^{B}\frac{n_b}{N}\left|\text{acc}(b)-\text{conf}(b)\right|.
\]
A model with ECE of 0 is perfectly calibrated in the sense that its predicted probabilities match observed outcome frequencies at every confidence level. Lower values of both the Brier score and ECE indicate better alignment between predicted probabilities and observed outcomes across the range of model output.
\subsection{Subgroup analysis inclusion criteria}
We attempted stratified evaluation by child sex, household wealth (quintile), urban/rural residence, and maternal education. A subgroup analysis in a country was included only if: (a) both outcome classes (``Anemia'' and ``no Anemia'') were present in that subgroup's data; and (b) at least 50 total observations remained after filtering. Subgroups failing these criteria (due to severe class imbalance or small sample after preprocessing) were excluded from that country's subgroup evaluation.

\subsection{Sensitivity analyses}
We conducted several sensitivity analyses to ensure robustness:
\begin{itemize}
  \item \textbf{Calibration bins:} We varied ECE binning schemes (equal-frequency vs.\ equal-width) and found no major change in calibration rankings.
  \item \textbf{Threshold choice:} We assessed model performance at fixed probability thresholds (e.g.\ 10--50\%) to test stability of accuracy and F1.
\end{itemize}

\section{With-Country Evaluation Result Table with CIs}
\label{app:within_country_eval}

\begin{table*}[t]
\caption{\textbf{Discriminative performance across countries.} 
Area under the receiver operating characteristic curve (AUC--ROC) is reported with 95\% bootstrap bias-corrected and accelerated (BCa) confidence intervals. Best-performing model within each country is shown in bold.}
\label{tab:auc_results}
\centering
\small
\begin{tabular}{lcccc}
\toprule
\textbf{Country} & \textbf{Logistic regression} & \textbf{LightGBM} & \textbf{XGBoost} & \textbf{TabPFN} \\
\midrule
Gabon (2019--2021) 
& 0.629 (0.590--0.667) 
& 0.620 (0.580--0.659) 
& 0.624 (0.585--0.663) 
& \textbf{0.645 (0.607--0.682)} \\

Malawi (2024) 
& \textbf{0.703 (0.665--0.737)} 
& 0.665 (0.626--0.703) 
& 0.685 (0.646--0.720) 
& 0.701 (0.665--0.737) \\

Rwanda (2019--2020) 
& \textbf{0.677 (0.634--0.718)} 
& 0.660 (0.614--0.702) 
& 0.648 (0.605--0.691) 
& 0.670 (0.624--0.712) \\

Sierra Leone (2019) 
& 0.644 (0.597--0.687) 
& 0.644 (0.595--0.688) 
& 0.650 (0.604--0.694) 
& \textbf{0.673 (0.625--0.713)} \\

Haiti (2016--2017) 
& 0.637 (0.600--0.672) 
& 0.646 (0.608--0.681) 
& 0.652 (0.615--0.687) 
& \textbf{0.665 (0.628--0.699)} \\

Jordan (2023) 
& 0.676 (0.638--0.713) 
& 0.657 (0.616--0.693) 
& 0.667 (0.627--0.704) 
& \textbf{0.683 (0.644--0.718)} \\

Armenia (2015--2016) 
& 0.721 (0.630--0.800) 
& 0.749 (0.653--0.823) 
& \textbf{0.764 (0.678--0.830)} 
& 0.757 (0.674--0.827) \\

Kyrgyz Republic (2012) 
& 0.687 (0.646--0.725) 
& 0.700 (0.660--0.738) 
& 0.714 (0.675--0.752) 
& \textbf{0.715 (0.677--0.754)} \\

Liberia (2019--2020) 
& 0.529 (0.462--0.592) 
& 0.596 (0.533--0.657) 
& \textbf{0.608 (0.546--0.667)} 
& 0.600 (0.538--0.662) \\

Bolivia (2008) 
& 0.677 (0.624--0.727) 
& 0.661 (0.607--0.711) 
& 0.684 (0.631--0.732) 
& \textbf{0.712 (0.662--0.758)} \\

Congo DRC (2021--2022) 
& 0.639 (0.613--0.664) 
& 0.623 (0.597--0.650) 
& 0.638 (0.611--0.664) 
& \textbf{0.652 (0.625--0.678)} \\

Guatemala (2014--2015) 
& 0.712 (0.686--0.735) 
& 0.716 (0.691--0.740) 
& 0.724 (0.698--0.746) 
& \textbf{0.730 (0.705--0.753)} \\

Ghana (2022) 
& 0.674 (0.636--0.710) 
& 0.663 (0.623--0.700) 
& 0.668 (0.628--0.705) 
& \textbf{0.680 (0.641--0.717)} \\

Gambia (2019--2020) 
& 0.697 (0.655--0.736) 
& 0.698 (0.656--0.738) 
& 0.702 (0.660--0.742) 
& \textbf{0.720 (0.680--0.759)} \\

Cameroon (2018) 
& 0.656 (0.617--0.694) 
& 0.667 (0.627--0.705) 
& 0.676 (0.637--0.713) 
& \textbf{0.680 (0.638--0.715)} \\

Mali (2023--2024) 
& 0.669 (0.633--0.702) 
& 0.668 (0.633--0.702) 
& 0.662 (0.626--0.696) 
& \textbf{0.684 (0.649--0.716)} \\
\bottomrule
\end{tabular}
\end{table*}

\section{LOCO Result Table with CIs}
\label{app:loco_table}
\begin{table*}[!htb]
\centering
\caption{Leave-One-Country-Out (LOCO) performance of Anemia prediction models reported as AUC-ROC with 95\% bootstrap BCa confidence intervals. Best performance per row is highlighted in bold.}
\label{tab:loco_auc_results}
\resizebox{\textwidth}{!}{
\begin{tabular}{lcccc}
\toprule
\textbf{Held-out country} & \textbf{LogReg} & \textbf{LightGBM} & \textbf{XGBoost} & \textbf{TabPFN} \\
\midrule

Gabon (2019--2021) & 0.630 (0.612--0.647) & \textbf{0.604 (0.586--0.622)} & 0.607 (0.590--0.624) & 0.608 (0.591--0.627) \\
Malawi (2024) & 0.673 (0.656--0.689) & 0.662 (0.644--0.679) & \textbf{0.667 (0.650--0.684)} & \textbf{0.667 (0.649--0.683)} \\
Rwanda (2019--2020) & 0.640 (0.620--0.660) & 0.644 (0.623--0.664) & 0.644 (0.624--0.663) & \textbf{0.649 (0.629--0.668)} \\
Sierra Leone (2019) & 0.620 (0.600--0.640) & 0.590 (0.570--0.610) & 0.592 (0.572--0.612) & \textbf{0.604 (0.582--0.624)} \\
Haiti (2016--2017) & 0.645 (0.628--0.662) & \textbf{0.645 (0.628--0.661)} & 0.635 (0.619--0.652) & 0.643 (0.627--0.659) \\
Jordan (2023) & 0.613 (0.594--0.630) & 0.580 (0.561--0.598) & 0.580 (0.562--0.599) & \textbf{0.601 (0.582--0.618)} \\
Armenia (2015--2016) & 0.634 (0.595--0.672) & 0.653 (0.611--0.695) & 0.641 (0.599--0.683) & \textbf{0.650 (0.606--0.688)} \\
Kyrgyz Republic (2012) & 0.635 (0.618--0.655) & 0.631 (0.612--0.650) & 0.631 (0.612--0.649) & \textbf{0.640 (0.621--0.659)} \\
Liberia (2019--2020) & 0.594 (0.568--0.622) & \textbf{0.584 (0.556--0.609)} & 0.576 (0.549--0.603) & 0.585 (0.557--0.612) \\
Bolivia (2008) & 0.495 (0.470--0.520) & \textbf{0.658 (0.636--0.680)} & 0.652 (0.629--0.675) & 0.638 (0.614--0.662) \\
Congo (DRC, 2021--2022) & 0.622 (0.610--0.634) & 0.587 (0.576--0.600) & \textbf{0.597 (0.584--0.609)} & 0.591 (0.579--0.604) \\
Guatemala (2014--2015) & 0.678 (0.668--0.690) & 0.688 (0.676--0.699) & \textbf{0.701 (0.690--0.712)} & 0.690 (0.679--0.701) \\
Ghana (2022) & 0.668 (0.651--0.685) & 0.569 (0.551--0.587) & \textbf{0.602 (0.584--0.621)} & 0.593 (0.575--0.612) \\
Gambia (2019--2020) & 0.674 (0.654--0.693) & 0.598 (0.578--0.617) & 0.623 (0.604--0.642) & \textbf{0.645 (0.626--0.664)} \\
Cameroon (2018) & 0.650 (0.632--0.667) & 0.613 (0.595--0.631) & \textbf{0.626 (0.608--0.644)} & 0.624 (0.606--0.641) \\
Mali (2023--2024) & 0.663 (0.647--0.678) & 0.631 (0.616--0.646) & 0.640 (0.624--0.655) & \textbf{0.645 (0.630--0.660)} \\
\bottomrule
\end{tabular}
}
\end{table*}

\section{Subgroup performance feasibility and data constraints}
Several country surveys had insufficient data to support subgroup analysis. In these cases, one or more strata (e.g.\ male vs.\ female) contained too few observations with a recorded Anemia outcome following age restriction and outcome exclusion to permit stable model evaluation. Armenia, Bolivia, and Jordan had no viable subgroup partitions under these conditions (see Table~\ref{tab:subgroup_counts}). This limitation reflects data sparsity inherent to the survey sample sizes in these settings relative to the subgroup granularity required, rather than an absence of subgroup variation in the underlying population.

\begin{footnotesize}
\setlength{\LTleft}{0pt}
\setlength{\LTright}{0pt}
\newlength{\cntrycol}\setlength{\cntrycol}{0.15\linewidth}
\newlength{\datacol}\setlength{\datacol}{0.164\linewidth}

\begin{longtable}{@{}p{\cntrycol}p{\datacol}p{\datacol}p{\datacol}p{\datacol}p{\datacol}@{}}

  \caption{%
    Sample sizes per demographic subgroup across all
    included DHS surveys. Each cell reports the number of observations
    remaining after outcome and covariate filtering for each stratum.
    $^\dagger$~Stratum excluded from ML analyses ($n < 50$ after
    filtering, or a single outcome class in the training split).
    \textit{n/a}~indicates the variable was absent from the survey.
  }
  \label{tab:subgroup_counts}\\

  \toprule
  \textbf{Country} & \textbf{Sex} & \textbf{Wealth} & \textbf{Education} & \textbf{Residence} & \textbf{Age group} \\
  \midrule
  \endfirsthead

  \multicolumn{6}{c}{\tablename\ \thetable{} -- \textit{continued}}\\[2pt]
  \toprule
  \textbf{Country} & \textbf{Sex} & \textbf{Wealth} & \textbf{Education} & \textbf{Residence} & \textbf{Age group} \\
  \midrule
  \endhead

  \midrule
  \multicolumn{6}{r}{\textit{Continued on next page}}\\
  \endfoot

  \bottomrule
  \endlastfoot

  \textbf{Armenia 2015--2016} & \multicolumn{5}{l}{\textit{All strata excluded from ML analyses}$^\dagger$} \\
  \textbf{Bolivia 2008}        & \multicolumn{5}{l}{\textit{All strata excluded from ML analyses}$^\dagger$} \\
  \textbf{Cameroon 2018} & male=976, female=941 & poorer=406, middle=450, poorest=263, richer=447, richest=351 & secondary=854, primary=659, no education=283, higher=121 & Urban=980, Rural=937 & 12--23m=701, 48--59m=171, 24--35m=436, 36--47m=288, 6--11m=264, 0--5m=57 \\
  \textbf{Congo DRC 2021--2022} & female=1,340, male=1,471 & poorest=736, poorer=719, middle=649, richer=432, richest=275 & secondary=1,393, primary=865, no education=468, higher=85 & Rural=2,017, Urban=794 & 12--23m=1,308, 24--35m=846, 6--11m=504, 0--5m=153, 36--47m=$^\dagger$, 48--59m=$^\dagger$ \\
  \textbf{Gabon 2019--2021} & female=1,201, male=1,256 & poorest=961, richer=320, middle=390, poorer=562, richest=224 & secondary=1,631, no education=134, primary=504, higher=188 & Urban=1,768, Rural=689 & 24--35m=573, 48--59m=281, 12--23m=834, 36--47m=372, 6--11m=319, 0--5m=78 \\
  \textbf{Gambia 2019--2020} & female=833, male=949 & middle=363, poorer=381, richer=271, richest=216, poorest=551 & no education=878, secondary=503, primary=335, higher=66 & Urban=874, Rural=908 & 24--35m=418, 36--47m=267, 6--11m=226, 48--59m=177, 12--23m=650, 0--5m=44$^\dagger$ \\
  \textbf{Ghana 2022} & female=898, male=911 & poorest=562, poorer=439, middle=340, richer=263, richest=205 & secondary=863, no education=524, primary=286, higher=136 & Rural=1,052, Urban=757 & 12--23m=878, 24--35m=537, 6--11m=314, 0--5m=80, 36--47m=$^\dagger$, 48--59m=$^\dagger$ \\
  \textbf{Guatemala 2014--2015} & female=3,169, male=3,472 & richest=1,001, richer=1,331, middle=1,397, poorer=1,414, poorest=1,498 & secondary=1,866, higher=290, primary=3,482, no education=1,003 & Urban=2,501, Rural=4,140 & 24--35m=1,637, 48--59m=934, 36--47m=1,286, 12--23m=1,969, 6--11m=708, 0--5m=107 \\
  \textbf{Haiti 2016--2017} & female=1,478, male=1,523 & poorer=644, middle=633, richest=423, richer=534, poorest=767 & primary=1,176, secondary=1,236, higher=107, no education=482 & Rural=2,021, Urban=980 & 24--35m=731, 12--23m=887, 36--47m=535, 48--59m=443, 6--11m=340, 0--5m=65 \\
  \textbf{Jordan 2023}         & \multicolumn{5}{l}{\textit{All strata excluded from ML analyses}$^\dagger$} \\
  \textbf{Kyrgyz Republic 2012} & male=1,129, female=1,050 & poorest=482, middle=491, richest=306, poorer=475, richer=425 & higher=964, secondary=1,208, primary=6$^\dagger$, no education=1$^\dagger$ & Urban=573, Rural=1,606 & 12--23m=687, 36--47m=386, 48--59m=266, 24--35m=472, 6--11m=313, 0--5m=55 \\
  \textbf{Liberia 2019--2020} & female=674, male=644 & middle=269, poorest=411, poorer=389, richer=155, richest=94 & secondary=371, primary=405, no education=524, higher=18$^\dagger$ & Rural=880, Urban=438 & 24--35m=300, 12--23m=431, 6--11m=168, 48--59m=154, 36--47m=218, 0--5m=47$^\dagger$ \\
  \textbf{Malawi 2024} & female=915, male=895 & poorer=347, richer=341, middle=332, richest=372, poorest=418, nan=---$^\dagger$ & primary=1,203, secondary=471, no education=102, higher=34$^\dagger$ & Rural=1,480, Urban=330 & 48--59m=---$^\dagger$, 24--35m=653, 12--23m=836, 6--11m=268, 36--47m=---$^\dagger$, 0--5m=53 \\
  \textbf{Mali 2023--2024} & female=1,047, male=1,036 & richest=427, richer=421, poorer=397, middle=463, poorest=375 & secondary=524, no education=1,115, primary=385, higher=59 & Urban=585, Rural=1,498 & 24--35m=677, 48--59m=---$^\dagger$, 12--23m=1,014, 36--47m=---$^\dagger$, 6--11m=301, 0--5m=91 \\
  \textbf{Rwanda 2019--2020} & male=1,165, female=1,116 & richer=430, richest=435, poorest=504, poorer=445, middle=467 & secondary=429, higher=110, primary=1,504, no education=238 & Rural=1,775, Urban=506 & 36--47m=421, 6--11m=249, 48--59m=265, 24--35m=599, 12--23m=691, 0--5m=56 \\
  \textbf{Sierra Leone 2019} & male=1,155, female=1,081 & poorest=567, middle=465, poorer=483, richer=430, richest=291 & no education=1,186, primary=345, secondary=637, higher=68 & Rural=1,485, Urban=751 & 24--35m=553, 12--23m=728, 48--59m=230, 36--47m=354, 6--11m=290, 0--5m=81 \\

\end{longtable}
\end{footnotesize}

\section{Model calibration across countries}
\begin{longtable}{llcccc}
\caption{Country-level calibration performance (Brier score and Expected Calibration Error). Rankings are computed within each country (lower is better).} \\
\label{tab:country_calibration} \\

\toprule
\textbf{Country} & \textbf{Model} & \textbf{Brier $\downarrow$} & \textbf{ECE $\downarrow$} & \textbf{Rank (ECE)} & \textbf{Rank (Brier)} \\
\midrule
\endfirsthead

\toprule
\textbf{Country} & \textbf{Model} & \textbf{Brier $\downarrow$} & \textbf{ECE $\downarrow$} & \textbf{Rank (ECE)} & \textbf{Rank (Brier)} \\
\midrule
\endhead

\midrule
\multicolumn{6}{r}{\textit{Continued on next page}} \\
\midrule
\endfoot

\bottomrule
\endlastfoot

\multirow{4}{*}{Armenia 2015--2016}
& \textbf{XGBoost} & \textbf{0.1189} & \textbf{0.0370} & \textbf{1} & \textbf{1} \\
& TabPFN & 0.1193 & 0.0508 & 2 & 2 \\
& LightGBM & 0.1270 & 0.0714 & 3 & 3 \\
& LogReg & 0.1358 & 0.0814 & 4 & 4 \\

\addlinespace

\multirow{4}{*}{Bolivia 2008}
& \textbf{TabPFN} & \textbf{0.2034} & \textbf{0.0278} & \textbf{1} & \textbf{1} \\
& XGBoost & 0.2107 & 0.0388 & 2 & 2 \\
& LogReg & 0.2223 & 0.0800 & 3 & 3 \\
& LightGBM & 0.2300 & 0.0945 & 4 & 4 \\

\addlinespace

\multirow{4}{*}{Cameroon 2018}
& LogReg & \textbf{0.2336} & \textbf{0.0363} & \textbf{1} & 2 \\
& \textbf{TabPFN} & \textbf{0.2322} & 0.0494 & 2 & \textbf{1} \\
& XGBoost & 0.2384 & 0.0610 & 3 & 3 \\
& LightGBM & 0.2453 & 0.0883 & 4 & 4 \\

\addlinespace

\multirow{4}{*}{Congo DRC 2021--2022}
& \textbf{TabPFN} & \textbf{0.2147} & \textbf{0.0198} & \textbf{1} & \textbf{1} \\
& LogReg & 0.2202 & 0.0281 & 2 & 4 \\
& XGBoost & 0.2182 & 0.0313 & 3 & 2 \\
& LightGBM & 0.2185 & 0.0368 & 4 & 3 \\

\addlinespace

\multirow{4}{*}{Gabon 2019--2021}
& \textbf{XGBoost} & \textbf{0.2048} & \textbf{0.0361} & \textbf{1} & \textbf{1} \\
& LogReg & 0.2074 & 0.0393 & 2 & 3 \\
& TabPFN & 0.2052 & 0.0526 & 3 & 2 \\
& LightGBM & 0.2107 & 0.0666 & 4 & 4 \\

\addlinespace

\multirow{4}{*}{Gambia 2019--2020}
& \textbf{LogReg} & \textbf{0.2240} & \textbf{0.0625} & \textbf{1} & 4 \\
& TabPFN & 0.2107 & 0.0644 & 2 & 1 \\
& LightGBM & 0.2207 & 0.0670 & 3 & 3 \\
& XGBoost & 0.2135 & 0.0724 & 4 & 2 \\

\addlinespace

\multirow{4}{*}{Ghana 2022}
& \textbf{TabPFN} & \textbf{0.2191} & \textbf{0.0256} & \textbf{1} & \textbf{1} \\
& LogReg & 0.2286 & 0.0467 & 2 & 3 \\
& XGBoost & 0.2256 & 0.0522 & 3 & 2 \\
& LightGBM & 0.2314 & 0.0591 & 4 & 4 \\

\addlinespace

\multirow{4}{*}{Guatemala 2014--2015}
& \textbf{TabPFN} & \textbf{0.1814} & \textbf{0.0233} & \textbf{1} & \textbf{1} \\
& LightGBM & 0.1851 & 0.0245 & 2 & 2 \\
& XGBoost & 0.1866 & 0.0343 & 3 & 3 \\
& LogReg & 0.1985 & 0.0497 & 4 & 4 \\

\addlinespace

\multirow{4}{*}{Haiti 2016--2017}
& \textbf{TabPFN} & \textbf{0.2057} & \textbf{0.0289} & \textbf{1} & \textbf{1} \\
& XGBoost & 0.2093 & 0.0433 & 2 & 2 \\
& LogReg & 0.2116 & 0.0537 & 3 & 3 \\
& LightGBM & 0.2119 & 0.0541 & 4 & 4 \\

\addlinespace

\multirow{4}{*}{Jordan 2023}
& \textbf{TabPFN} & \textbf{0.2099} & \textbf{0.0367} & \textbf{1} & \textbf{1} \\
& XGBoost & 0.2135 & 0.0409 & 2 & 2 \\
& LogReg & 0.2172 & 0.0539 & 3 & 3 \\
& LightGBM & 0.2180 & 0.0691 & 4 & 4 \\

\addlinespace

\multirow{4}{*}{Kyrgyz Republic 2012}
& \textbf{LogReg} & \textbf{0.2374} & \textbf{0.0374} & \textbf{1} & 3 \\
& TabPFN & 0.2279 & 0.0505 & 2 & 1 \\
& XGBoost & 0.2315 & 0.0576 & 3 & 2 \\
& LightGBM & 0.2433 & 0.0907 & 4 & 4 \\

\addlinespace

\multirow{4}{*}{Liberia 2019--2020}
& \textbf{TabPFN} & \textbf{0.1991} & \textbf{0.0631} & \textbf{1} & \textbf{1} \\
& LogReg & 0.2048 & 0.0634 & 2 & 2 \\
& XGBoost & 0.2053 & 0.0704 & 3 & 3 \\
& LightGBM & 0.2179 & 0.1154 & 4 & 4 \\

\addlinespace

\multirow{4}{*}{Malawi 2024}
& \textbf{TabPFN} & \textbf{0.2258} & \textbf{0.0366} & \textbf{1} & \textbf{1} \\
& XGBoost & 0.2272 & 0.0389 & 2 & 2 \\
& LogReg & 0.2347 & 0.0601 & 3 & 4 \\
& LightGBM & 0.2302 & 0.0628 & 4 & 3 \\

\addlinespace

\multirow{4}{*}{Mali 2023--2024}
& \textbf{XGBoost} & \textbf{0.1809} & \textbf{0.0284} & \textbf{1} & 2 \\
& TabPFN & 0.1790 & 0.0327 & 2 & 1 \\
& LogReg & 0.1854 & 0.0437 & 3 & 4 \\
& LightGBM & 0.1819 & 0.0547 & 4 & 3 \\

\addlinespace

\multirow{4}{*}{Rwanda 2019--2020}
& \textbf{TabPFN} & \textbf{0.2091} & \textbf{0.0438} & \textbf{1} & \textbf{1} \\
& XGBoost & 0.2119 & 0.0539 & 2 & 3 \\
& LogReg & 0.2222 & 0.0627 & 3 & 4 \\
& LightGBM & 0.2118 & 0.0629 & 4 & 2 \\

\addlinespace

\multirow{4}{*}{Sierra Leone 2019}
& \textbf{LogReg} & \textbf{0.2054} & \textbf{0.0342} & \textbf{1} & 2 \\
& TabPFN & 0.2043 & 0.0615 & 2 & 1 \\
& XGBoost & 0.2073 & 0.0665 & 3 & 3 \\
& LightGBM & 0.2171 & 0.0906 & 4 & 4 \\

\addlinespace

\end{longtable}

\section{Interpretability Analysis For Other Models}
\label{app:other_model_interpretability}
\begin{figure}[H]
\centering
\includegraphics[width=\textwidth]{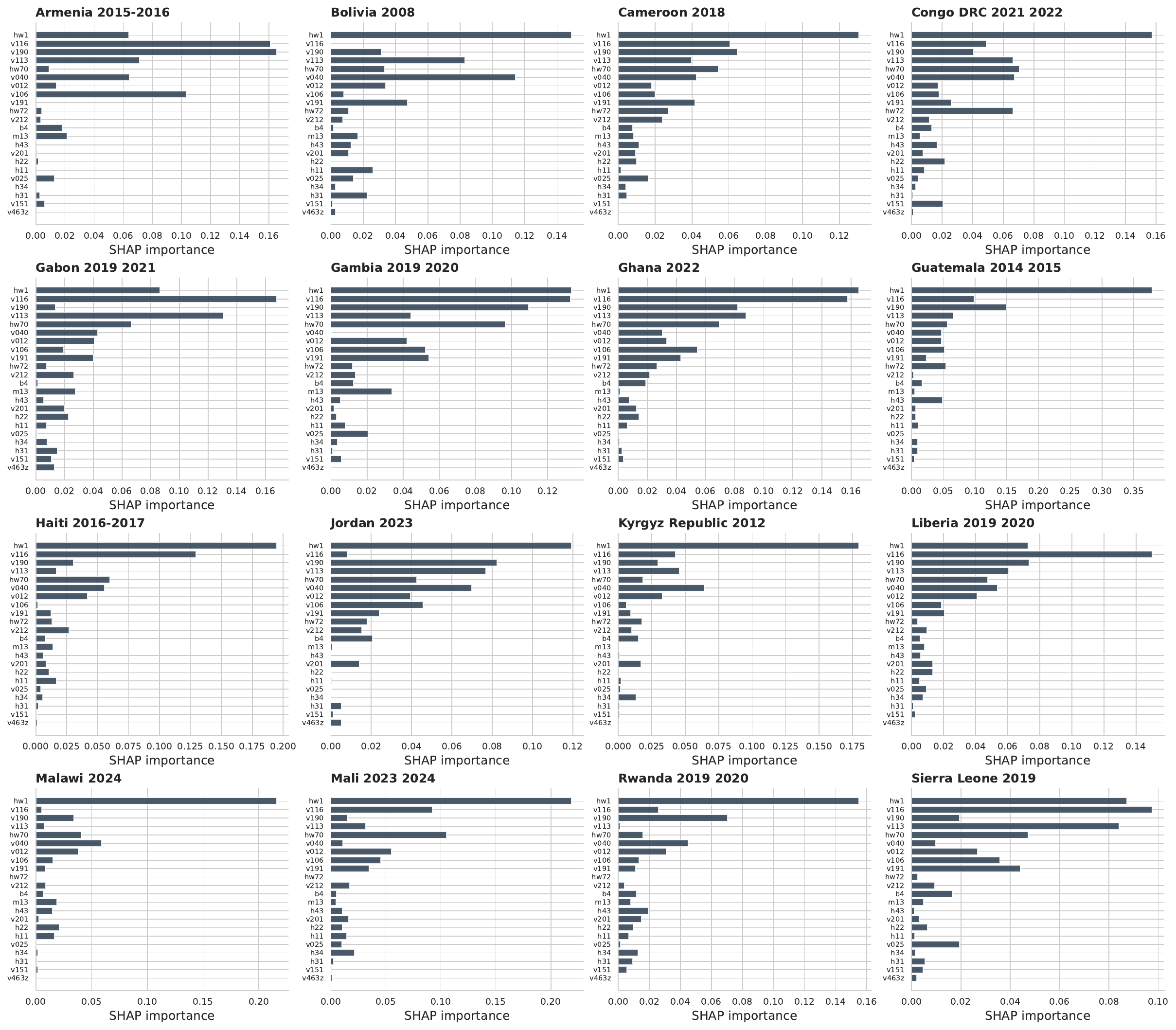}
\caption{\textbf{Country-level SHAP feature importance for logistic regression.} Each panel shows SHAP values for each predictor within a single country, estimated on the held-out test set under within country evaluation. Child age (\texttt{hw1}) was the dominant predictor across nearly all settings, followed by altitude (\texttt{v040}) and anthropometric z-scores (\texttt{hw70}, \texttt{hw72}). The relative contribution of socioeconomic indicators (\texttt{v191}, \texttt{v106}) and illness-related variables varied across countries, reflecting heterogeneity in the epidemiological profile of Anemia across the included populations.}
\label{fig:LogReg_feature_importance_countries}
\end{figure}

\begin{figure}[H]
\centering
\includegraphics[width=\textwidth]{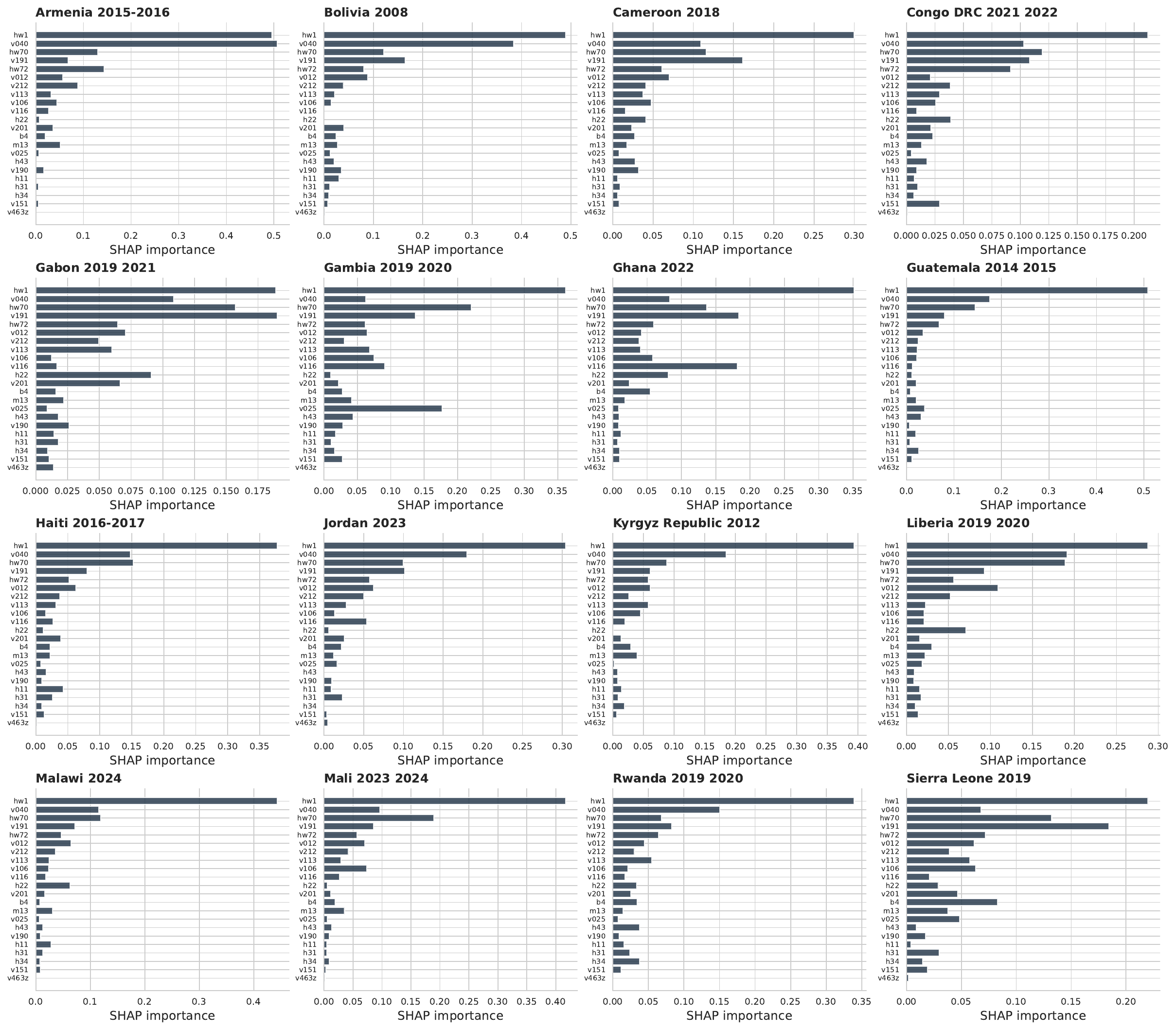}
\caption{\textbf{Country-level SHAP feature importance for XGBoost.} Each panel shows SHAP values for each predictor within a single country, estimated on the held-out test set under within country evaluation. Child age (\texttt{hw1}) and altitude (\texttt{v040}) were the dominant predictors across most settings, with anthropometric z-scores (\texttt{hw70}, \texttt{hw72}) and wealth index (\texttt{v191}) contributing consistently in the upper tier. The relative ranking of socioeconomic, maternal, and illness-related features varied across countries, reflecting the heterogeneity of Anemia determinants across the included populations.}
\label{fig:XGBoost_feature_importance_countries}
\end{figure}

\begin{figure}[H]
\centering
\includegraphics[width=\textwidth]{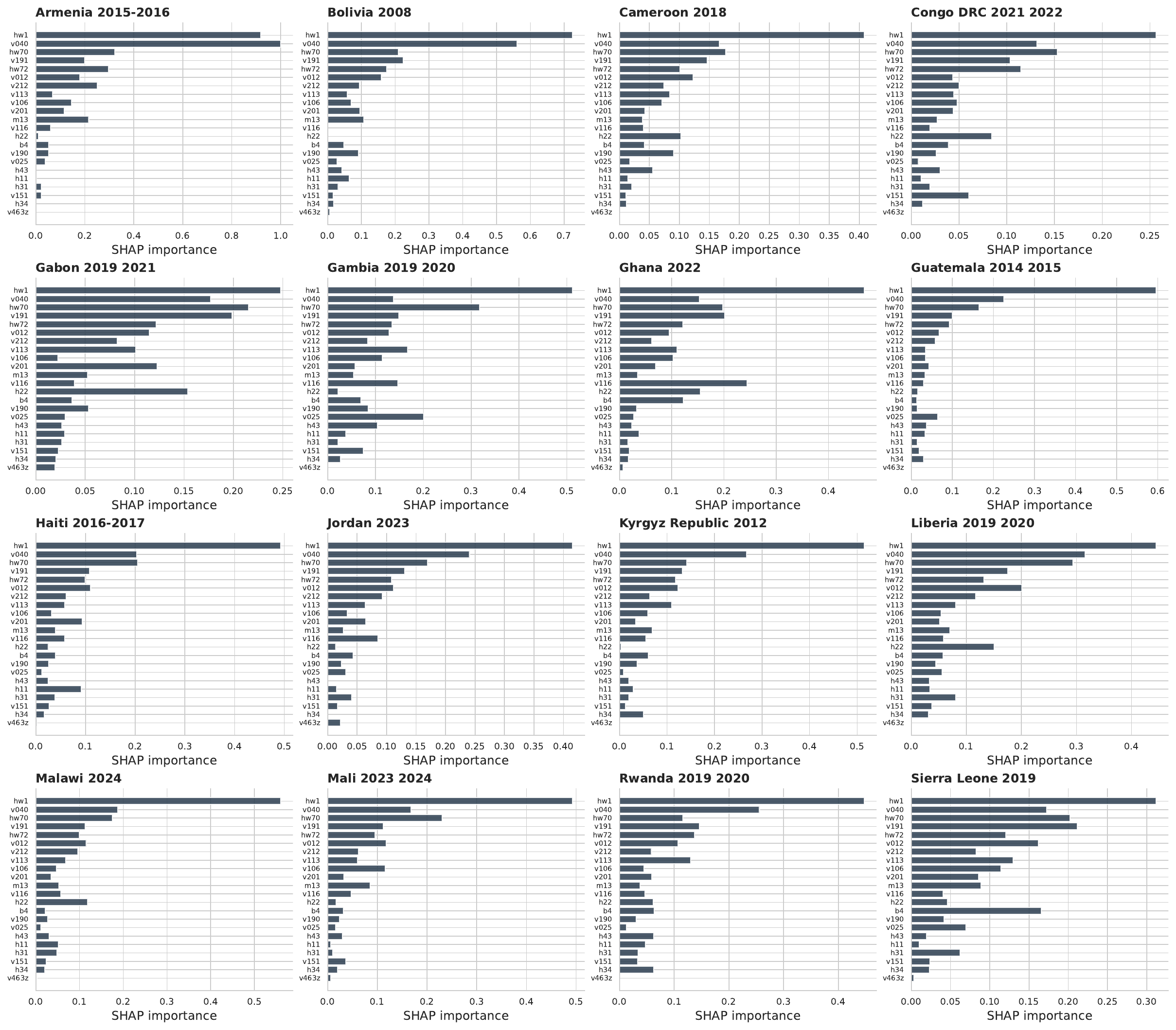}
\caption{\textbf{Country-level SHAP feature importance for LightGBM.} Each panel shows SHAP values for each predictor within a single country, estimated on the held-out test set under within country evaluation. Child age (\texttt{hw1}) was the dominant predictor across all settings, with notably large absolute SHAP values relative to the other models, particularly in Rwanda, Malawi, Haiti, Guatemala, Cameroon, Armenia and Bolivia. Altitude (\texttt{v040}) ranked consistently in the upper tier, followed by anthropometric z-scores (\texttt{hw70}, \texttt{hw72}) and wealth index (\texttt{v191}). The long tail of lower-ranked features, including illness indicators, deworming, and smoking status, showed greater cross-country variability in their relative contributions than the top-ranked predictors.}
\label{fig:LightGBM_feature_importance_countries}
\end{figure}

\section{Reverse LOCO Train-Test Performance Network}
\label{app:reverse_loco_network}
\begin{figure}[H]
\centering
\includegraphics[width=\textwidth]{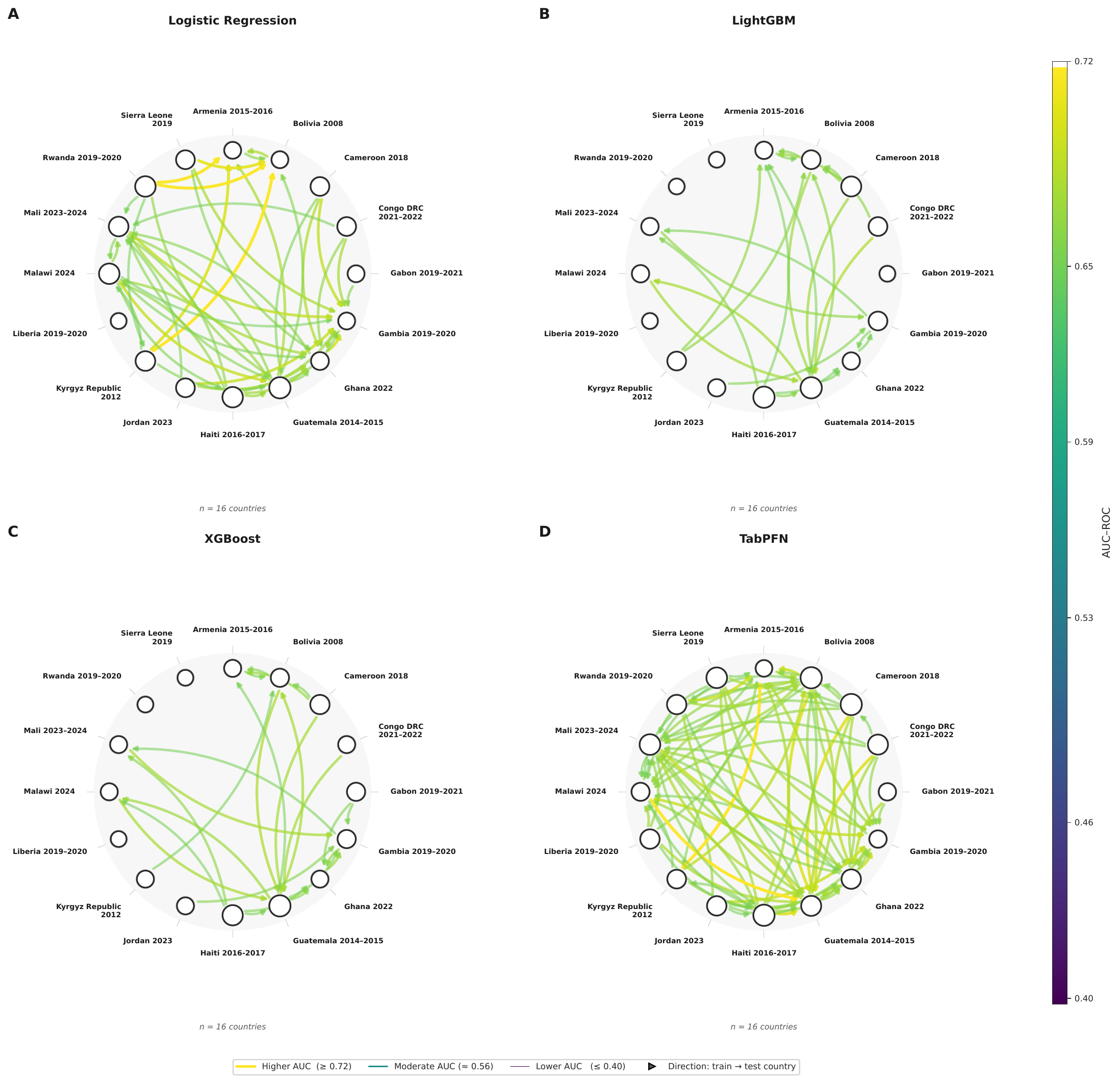}
\caption{\textbf{Cross-country transferability of predictive models under reverse LOCO validation, displayed as directed performance networks.} Each panel shows AUC-ROC for a given model (A: logistic regression; B: LightGBM; C: XGBoost; D: TabPFN v2.6) across all 240 directed train--test country pairs, arranged as a circular network of the 16 study countries. Each directed edge connects a training source country to a held-out test country, with edge color encoding AUC-ROC on a shared scale (yellow $\geq 0.72$; grey $\approx 0.56$; dark blue $\leq 0.40$). Node size reflects the mean AUC-ROC achieved when that country serves as a training source across all target countries, providing a summary of each population's utility as a generalizable training domain. Denser concentrations of high-AUC edges originating from a node indicate epidemiologically informative source populations; column-wise clustering of low-AUC incoming edges indicates target populations that are persistently difficult to predict regardless of training source or model. The structural pattern of asymmetric transferability is broadly consistent across all four models, with Ghana, Guatemala, Haiti, and Bolivia emerging as comparatively stronger source domains and Armenia and Liberia as consistently harder target populations.}
\label{fig:reverse_loco_network_plot}
\end{figure}
\end{document}